\documentclass{article}
\usepackage[accepted]{icml}

\usepackage{url}
\definecolor{citecolor}{RGB}{17,80,197}
\definecolor{linkcolor}{HTML}{ED1C24}
\usepackage[breaklinks=true,colorlinks,citecolor=citecolor,linkcolor=linkcolor,bookmarks=false]{hyperref}

\usepackage{placeins}
\usepackage{newtxtext}
\usepackage{array}
\usepackage{pifont}
\usepackage{tabularx}
\usepackage{adjustbox}
\usepackage{multirow}
\usepackage{enumitem}
\usepackage{xspace}
\usepackage{booktabs,dcolumn}
\usepackage{amsmath,amsthm,amsfonts,amssymb,bm,bbm}
\usepackage[noorphans,vskip=0.75ex,leftmargin=2ex]{quoting}
\usepackage{float}
\usepackage{vcell}
\usepackage{colortbl}
\usepackage{bbding}
\usepackage{wrapfig}
\usepackage{adjustbox}
\usepackage{pifont}
\usepackage{diagbox}
\usepackage{makecell}
\usepackage{tcolorbox}
\usepackage{xcolor}
\usepackage{listings}
\usepackage{sourcecodepro}
\usepackage{stmaryrd}
\usepackage[utf8]{inputenc}  % Ensure UTF-8 encoding
\makeatletter
\lst@InstallKeywords k{attributes}{attributestyle}\slshape{attributestyle}{}ld
\makeatother

\definecolor{pythonblue}{rgb}{0.16,0.12,0.93}
\definecolor{cppgreen}{rgb}{0.16,0.42,0.16}
\definecolor{promptinsert}{HTML}{bfefff}
\definecolor{compcolor}{HTML}{90EE90}
\definecolor{codehlcolor}{HTML}{ffec8b}
\definecolor{codehlcolor2}{HTML}{ffbbff}
\definecolor{bgcolor}{rgb}{0.95,0.95,0.92}
\definecolor{spblue}{HTML}{00b5ea}

\lstdefinestyle{python}{
    language=Python,
    basicstyle=\fontsize{8}{10}\ttfamily,
    keywordstyle=\color{blue},
    commentstyle=\color{gray},
    stringstyle=\color{black},
    showstringspaces=false,
    breaklines=true,
    breakindent=0pt,
    breakatwhitespace=false,
    escapeinside={(*@}{@*)}
}

\lstdefinestyle{cpp}{
    language=C++,
    basicstyle=\fontsize{8}{10}\ttfamily,
    keywordstyle=\color{blue},
    commentstyle=\color{gray},
    stringstyle=\color{green},
    showstringspaces=false,
    breaklines=true,
    breakindent=0pt,
    breakatwhitespace=false,
    escapeinside={(*@}{@*)}
}

\lstdefinestyle{plain}{
    basicstyle=\fontsize{8}{10}\ttfamily,
    keywordstyle=\color{blue},
    commentstyle=\color{gray},
    stringstyle=\color{green},
    showstringspaces=false,
    breaklines=true,
    breakatwhitespace=false,
    breakindent=0pt,
    escapeinside={(*@}{@*)},
    literate={á}{{\'a}}1 {ã}{{\~a}}1 {é}{{\'e}}1,
}

\lstdefinestyle{demo}{
    basicstyle=\fontsize{7}{8}\ttfamily,
    keywordstyle=\color{blue},
    commentstyle=\color{gray},
    stringstyle=\color{green},
    showstringspaces=false,
    breaklines=true,
    breakatwhitespace=false,
    breakindent=0pt,
    escapeinside={(*@}{@*)},
    literate={á}{{\'a}}1 {ã}{{\~a}}1 {é}{{\'e}}1,
}

\lstdefinestyle{example}{
    basicstyle=\fontsize{8}{10}\ttfamily,
    keywordstyle=\color{spblue}\bfseries\underline,
    commentstyle=\color{gray},
    stringstyle=\color{green},
    showstringspaces=false,
    breaklines=true,
    breakatwhitespace=false,
    breakindent=0pt,
    escapeinside={(*@}{@*)},
    morekeywords={ Question, Answer, Prediction, Results, Explanation },
}

\lstdefinestyle{python2}{
    language=Python,
    basicstyle=\fontsize{8}{10}\ttfamily,
    keywordstyle=\color{blue},
    commentstyle=\color{gray},
    stringstyle=\color{green},
    showstringspaces=false,
    breakatwhitespace=false,
    breaklines=true,
    breakindent=0pt,
    escapeinside={(*@}{@*)}
}

\lstdefinestyle{cpp2}{
    language=C++,
    basicstyle=\fontsize{8}{10}\ttfamily,
    keywordstyle=\color{blue},
    commentstyle=\color{gray},
    stringstyle=\color{green},
    showstringspaces=false,
    breaklines=true,
    breakindent=0pt,
    breakatwhitespace=false,
    escapeinside={(*@}{@*)}
}

\lstdefinestyle{sql}{
    language=SQL,
    basicstyle=\fontsize{8}{10}\ttfamily,
    keywordstyle=\color{blue},
    commentstyle=\color{green},
    stringstyle=\color{black},
    showstringspaces=false,
    breakatwhitespace=false,
    breaklines=true,
    breakindent=0pt,
    escapeinside={(*@}{@*)}
}

\lstdefinestyle{prompt}{
    language=Python,
    basicstyle=\fontsize{8}{10}\ttfamily,
    keywordstyle=\color{blue},
    commentstyle=\color{gray},
    showstringspaces=false,
    breaklines=true,
    keepspaces=true, 
    breakindent=0pt,
    breakatwhitespace=false,
    showspaces=false,   
    escapeinside={(*@}{@*)}
}
\lstdefinestyle{text}{
    basicstyle=\fontsize{8}{10}\ttfamily,
    showstringspaces=false,
    breaklines=true,
    breakatwhitespace=false,
    breakindent=0pt,
    keepspaces=true,
    showspaces=false,   
    escapeinside={(*@}{@*)}
}

\theoremstyle{plain}

\theoremstyle{definition}

\theoremstyle{remark}

\definecolor{mygreen}{RGB}{0, 128, 0}
\definecolor{myred}{RGB}{255, 0, 0}
\newcommand{\increase}[1]{\textcolor{mygreen}{+#1}}
\newcommand{\decrease}[1]{\textcolor{myred}{-#1}}

\newcommand{\openrag}{\textsc{Open-Rag}\xspace}
\newcommand{\sidr}{\textsc{SiDR}\xspace}
\newcommand{\dpr}{\textsc{DPR}\xspace}
\newcommand{\replug}{\textsc{RePlug}\xspace}
\newcommand{\selfrag}{\textsc{Self-RAG}\xspace}
\newcommand{\efive}{$\textsc{E5}$\xspace}
\newcommand{\contriever}{$\textsc{Contriever}$\xspace}
\newcommand{\contrieverms}{$\textsc{Contriever}_\textsc{MS}$\xspace}
\newcommand{\sidrms}{$\textsc{SiDR}_\textsc{MS}$\xspace}
\newcommand{\sidrnq}{$\textsc{SiDR}_\textsc{NQ}$\xspace}
\newcommand{\sidrtqa}{$\textsc{SiDR}_\textsc{TQA}$\xspace}
\newcommand{\jiawei}[1]{}

\begin{document}

\twocolumn[
% \icmltitle{Consistent-RAG: Learning In-Context Relevance End-to-end towards RAG}

% \icmltitle{Optimizing RAG End-to-End through In-Context Retrieval Learning}

% \icmltitle{Learning In-Context Relevance End-to-End towards RAG}

\icmltitle{OpenRAG: Optimizing RAG End-to-End via In-Context Retrieval Learning}

\begin{icmlauthorlist}
\icmlauthor{Jiawei Zhou}{hkust}
\icmlauthor{Lei Chen}{hkust} \\
$^1$Hong Kong University of Science and Technology
\end{icmlauthorlist}

\icmlkeywords{Machine Learning, ICML}

\vskip 0.3in
]

% \printAffiliationsAndNotice{}

\begin{abstract}
% Despite widespread adoption, existing retrieval-augmented generation (RAG) frameworks commonly employ off-the-shelf retrievers with large language models (LLMs) without further joint training. 
In this paper, we analyze and empirically show that the learned relevance for conventional information retrieval (IR) scenarios may be inconsistent in retrieval-augmented generation (RAG) scenarios. To bridge this gap, we introduce \textbf{\openrag}, a RAG framework that is \underline{\textbf{OP}}timized \underline{\textbf{EN}}d-to-end by tuning the retriever to capture in-context relevance, enabling adaptation to the diverse and evolving needs. Extensive experiments across a wide range of tasks demonstrate that \openrag, by tuning a retriever end-to-end, leads to a consistent improvement of 4.0\% over the original retriever, consistently outperforming existing state-of-the-art retrievers by 2.1\%. Additionally, our results indicate that for some tasks, an end-to-end tuned 0.2B retriever can achieve improvements that surpass those of RAG-oriented or instruction-tuned 8B large language models (LLMs), highlighting the cost-effectiveness of our approach in enhancing RAG systems.
\end{abstract}
\section{Introduction}
As large language models (LLMs)~\cite{zhao2023survey,minaee2024large} scale, they face a data bottleneck where the high-quality internet data unable to meet growing training demands. Meanwhile, the volume of downstream data is expanding rapidly but often remains unusable for pre-training due to their real-time availability~\cite{wang2024survey, liu2023fingpt}, privacy concerns~\cite{arora2023knowledge}, licensing restrictions~\cite{min2024silo}, and ethical concern~\cite{serouis2024exploring, ayyamperumal2024current}.

Retrieval-augmented generation (RAG)~\cite{lewis2020retrieval, guu2020retrieval, gao2023retrieval} emerges as a promising solution to this challenge. Rather than relying solely on well-curated internet data, RAG leverages information retrieval (IR) to fetch relevant data from external sources and incorporates it as context to enhance generation quality. This is valuable as RAG enables the use of rapidly expanding yet often inaccessible downstream data, which are more scalable and up-to-date than the heavily processed and regulated internet data used in pre-training.

% Retrieval-augmented generation (RAG)~\cite{lewis2020retrieval, guu2020retrieval, gao2023retrieval} is a widely used framework that combines the generative capabilities of large language models (LLMs)~\cite{zhao2023survey} with information retrieval (IR)~\cite{manning2009introduction} to access external knowledge. This approach effectively addresses challenges that standalone LLMs cannot handle, such as managing long-tail knowledge~\cite{mallen2023not, soudani2024fine} and private data~\cite{arora2023knowledge, min2024silo}, providing up-to-date information~\cite{wang2024survey, liu2023fingpt}, and adapting to specific domains and tasks~\cite{jeong2024adaptive, xiong2024benchmarking} without additional tuning.

\begin{figure}[t]
\begin{center}
\includegraphics[width=0.98\columnwidth]{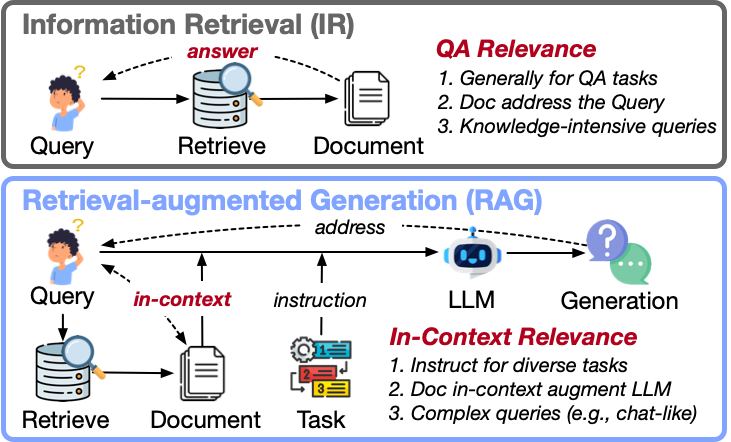}
\vspace{-0.1cm}
\caption{Comparison of query-document relevance in IR scenario and RAG scenario.}
\label{fig:fig1}
\end{center}
\vspace{-0.6cm}
\end{figure}

Despite their success, existing RAG frameworks typically rely on off-the-shelf retrievers trained on QA datasets, which can lead to inconsistencies between the learned retrieval relevance and the needs of downstream tasks. This discrepancy highlights key relevance gaps between IR and RAG scenarios. We explore these gaps in detail below, drawing on insights from prior research. 
First, there is the \textbf{broadening of tasks}: traditional IR datasets~\cite{kwiatkowski2019natural, bajaj2016ms} are designed mainly for open-domain question-answering (OpenQA), while RAG framework are applied to a wider range of tasks, such as recommendation~\cite{manzoor2022towards}, dialog systems~\cite{liu2024chatqa}, and role-playing~\cite{wang2023does}, where task requirements can be flexibly written as instructions. We refer to relevance in these two cases as \emph{QA relevance} and \emph{in-context relevance}, respectively, as shown in Figure~\ref{fig:fig1}.
Second, the \textbf{role of retrieved documents} has shifted: in IR, retrieved documents are the final output provided to users, whereas in RAG, they are fed into the LLM to generate a response. Recent studies~\cite{cuconasu2024power, cuconasu2024rethinking, wu2024easily} have shown that including more answer-containing documents, which align with QA relevance in IR scenarios, can harm RAG performance, while documents without direct answers may actually help. These findings challenge traditional IR assumptions in the RAG setting.
Finally, the \textbf{complexity of queries} has increased: unlike traditional IR, where queries are typically simple questions, RAG queries tend to be more diverse and noisy, reflecting varying levels of task complexity. Several studies highlight the challenges of complex queries and suggest that refining queries~\cite{chan2024rq} or generating task-specific queries~\cite{wu2024llm, koo2024optimizing} based on documents can significantly enhance RAG performance. 

To address this gap, we introduce \textbf{\openrag}, a RAG framework that is \underline{\textbf{OP}}timized \underline{\textbf{EN}}d-to-end by tuning the retriever to capture in-context relevance. Unlike existing retrievers, which are constrained to training on specific corpora and tasks with human annotations provided, our framework is \textbf{OPEN} to training on any task, with any corpus and any LLM. During training, \openrag retrieves documents on-the-fly and identifies them as positives or negatives for contrastive learning.To reduce training costs, we use approximation techniques to bypass the autoregressive generation process and employ semi-parametric retrieval to avoid the need for re-indexing. Our training requires only four GPUs and can be completed within a day. Extensive experiments demonstrate that our method leads to significant improvements, consistently outperforming state-of-the-art (SOTA) retrievers. For certain tasks, our improvements surpass those achieved by tuning an 8B LLM, showcasing that end-to-end retrieval learning is a cost-effective approach for enhancing RAG systems.

Our contribution can be summarized as follows:

$\bullet$ We investigate the relevance gap between IR and RAG scenarios, providing empirical evidence of when and how  this gap negatively impacts RAG performance. 

$\bullet$ Through our experiments, we identify potential biases in prior research that may impede progress in this field. These findings provide critical insights to guide future research directions.

$\bullet$ We introduce \openrag, an end-to-end optimized RAG framework that learns in-context retrieval for various downstream tasks without requiring query-document human annotations, facilitating broader real-world deployment and applications.

$\bullet$ Extensive experiments show that \openrag achieves superior performance across diverse tasks compared to RAG systems using SOTA retrievers or fine-tuned LLMs, underscoring its effectiveness as a reliable and versatile solution for improving RAG systems.

\section{Preliminary}
\subsection{Transferring from IR to RAG Scenarios}
\label{sec:ir_to_rag}

In Table~\ref{tab:preliminary}, we examine the performance of off-the-shelf retrievers across different datasets in IR and RAG scenarios. Details about the datasets and retrievers can be found in Appendix~\ref{appendix:dataset} and~\ref{appendix:models}, while the evaluation metric is described in Section~\ref{sec:setup}. Key findings are summarized below.

\begin{table}[ht]\small
\centering
\small
\caption{Accuracy in IR and RAG scenarios using Llama3-8b with top-1 retrieved document in-context; \textbf{Bold}: best performance; $\Delta$: \textcolor{mygreen}{improvement} or \textcolor{myred}{decline} compared to \sidrms; $\S$: has accessed the training split of the dataset.}
\resizebox{0.99\linewidth}{!}{
    \setlength{\tabcolsep}{0.5mm}{
\begin{tabular}{@{}cc>{\columncolor{gray!20}}cc>{\columncolor{gray!20}}cc>{\columncolor{gray!20}}cc>{\columncolor{gray!20}}cc>{\columncolor{gray!20}}cc>{\columncolor{gray!20}}c@{}}
% Table generated by Excel2LaTeX from sheet 'prelimnary-1doc (color)'
\toprule
\textbf{Dataset ($\rightarrow$)} & \multicolumn{4}{c}{\textbf{NQ}} & \multicolumn{4}{c}{\textbf{TriviaQA}} & \multicolumn{2}{c}{\textbf{PubHealth}} & \multicolumn{2}{c}{\textbf{ARC-C}} \\
   \cmidrule(lr){2-5} \cmidrule(lr){6-9} \cmidrule(lr){10-11} \cmidrule(lr){12-13}
\textbf{Retriever ($\downarrow$)} & \textit{\textbf{IR}} & $\Delta$ & \textit{\textbf{RAG}} & $\Delta$ & \textit{\textbf{IR}} & $\Delta$ & \textit{\textbf{RAG}} & $\Delta$ & \textit{\textbf{RAG}} & $\Delta$ & \textit{\textbf{RAG}} & $\Delta$ \\
\midrule
\multicolumn{13}{l}{\textit{\textbf{Unsupervised Pre-training}}} \\
Contriever & 23.6 &  \decrease{15.5} & 30.9 &  \decrease{3.5} & 37.2 &  \decrease{18.9} & 56.6 &  \decrease{5.4} & 61.8 & \decrease{1.7} & \textbf{58.6} & \textbf{\increase{1.7}} \\
E5-unsup & 30.8 & \decrease{8.3} & 33.4 &  \decrease{1.0} & 39.5 &   \decrease{16.6} & 54.3 &  \decrease{7.7} & 62.9 & \decrease{0.6} & 58.3 & \increase{1.4} \\
\midrule
\multicolumn{13}{l}{\textit{\textbf{Supervised on MSMARCO}}} \\
\quad DPR$_\text{MS}$ & 38.9 &  \decrease{0.2} & 34.9 &  \increase{0.5} & 43.7 &   \decrease{12.4} & 55.2 &  \decrease{6.8} & 64.5 &  \increase{1.0} & 56.3 & \decrease{0.6} \\
\quad SiDR$_\text{MS}$ & 39.1 &  -- & 34.4 &  -- & 56.1 &  -- & 62.0 &  -- & 63.5 & --  & 56.9 & -- \\
\multicolumn{13}{l}{\textit{\textbf{Supervised on NQ}}} \\
\quad DPR$_\text{NQ}$ & $\ddagger$43.5 &  \increase{4.4} & $\ddagger$38.5 &  \increase{4.1} & 39.4 &   \decrease{16.7} & 55.9 &  \decrease{6.1} & 62.9 & \decrease{0.6} & 56.6 & \decrease{0.3} \\
\quad SiDR$_\text{NQ}$ & $\ddagger$49.5 &  \increase{10.4} & $\ddagger$42.7 &  \increase{8.3} & 47.4 &  \decrease{8.7} & 59.8 &    \decrease{2.2} & 63.5 & --  & 57.1 & \increase{0.2} \\
\multicolumn{13}{l}{\textit{\textbf{Supervised on TQA}}} \\
\quad DPR$_\text{TQA}$ & 32.1 &  \decrease{7.0} & 32.9 &  \decrease{1.5} & $\ddagger$55.4 &  \decrease{0.7} & $\ddagger$61.1 &  \decrease{0.9} & 63.1 & \decrease{0.4} & 56.7 & \decrease{0.2} \\
\quad SiDR$_\text{TQA}$ & 30.6 &  \decrease{8.5} & 32.9 &  \decrease{1.5} & $\ddagger$56.9 &  \increase{0.8} & $\ddagger$\textbf{63.6} &  \textbf{ \increase{1.6}} & 61.1 &  \decrease{2.4} & \textbf{58.6} & \textbf{\increase{1.7}} \\
\midrule
\multicolumn{13}{l}{\textit{\textbf{Pre-training + Supervised on Multiple Datasets}}} \\
\quad Contriever$_\text{MS}$ & 41.5 &  \increase{2.4} & 36.5 &  \increase{2.1} & 53.5 &  \decrease{2.6} & 60.7 &   \decrease{1.3} & 63.1 & \decrease{0.4} & 58.1 & \increase{1.2} \\
\quad \efive & $\ddagger$\textbf{58.0} &  \textbf{\increase{18.9}} & $\ddagger$\textbf{43.2} &  \textbf{\increase{8.8}} & \textbf{58.7} &  \textbf{ \increase{2.6}} & 63.2 &   \increase{1.2} & \textbf{64.7} & \textbf{ \increase{1.2}} & 58.0 & \increase{1.1} \\
\midrule
\multicolumn{13}{l}{\textit{\textbf{Potential Improvement of IR vs. Improvement of LLMs}}} \\
\quad Best-of-8 & -- &  --  & 77.6 &  --  & -- &  --  & 80.3 &  --  & 92.1 &  --  & 71.5 & -- \\
\quad \efive + 8B-Instruct & -- &  --  & 54.4 &  --  & -- &  --  & 66.7 &  --  & 72.4 &  --  & 74.1 & -- \\
\quad \efive + 70B & -- &  --  & 51.4 &  --  & -- &  --  & 68.0 &  --  & 63.2 &  --  & 81.9 & -- \\
\bottomrule
\end{tabular}%
}}
\vspace{-0.8cm}
\label{tab:preliminary}
\end{table}

\textbf{Finding 1: Training retrievers in-domain is effective for both IR and RAG.}
As shown, with comparable training complexity, \sidrnq excels on the NQ dataset relative to other \sidr and \dpr models. Additionally, \sidrtqa outperforms the state-of-the-art retriever \efive in RAG scenarios on the TriviaQA dataset.

\textbf{Finding 2: Superiority of retrievers in IR scenarios can transfer to RAG scenarios cross-domain but not cross-task.}
For QA tasks, retrievers with higher accuracy in IR scenarios tends to perform better in RAG scenarios, as evidenced by NQ and TQA datasets. However, this trend does not extend to non-QA tasks. For instance, on the PubHealth dataset, the relatively weaker retriever DPR$_\text{MS}$ outperforms others, while on the ARC dataset, the unsupervised retriever \contriever surpasses all advanced retrievers.

\textbf{Finding 3: Retrieval has great potential to improve RAG as much as using instruction-tuned or larger LLMs.}
We use the \textit{Best-of-8} metric to measure the proportion of queries that can be addressed in RAG scenarios by any of the above eight retrievers. \textit{Best-of-8} substantially outperforms SOTA retriever E5 across these datasets. Notably, for most tasks, it even surpasses the combination of E5 with instruction-tuned LLMs (Llama3-8B-Instruct) or larger LLMs (Llama3-70B). For example, on NQ dataset, 77\% of test queries have a searchable document in the datastore that can serve as context to generate a correct answer. However, combining E5 with instruction-tuned LLMs addresses 54\% while larger LLMs address 51\%. These results highlight the largely untapped potential of million-scale datastores and in-context examples for enhancing LLM inference, where a well-optimized retrieval model could unlock this potential.

Motivated by these observations, our work aims to learns task-specific in-context relevance for RAG in an end-to-end manner, moving beyond the traditional QA relevance.

\subsection{Problem Setup}
\label{sec:problem}

\newcommand{\ir}{\ensuremath{\mathcal{R}_\theta}\xspace}
\newcommand{\llm}{\ensuremath{\mathcal{G}_\phi}\xspace}
\newcommand{\datastore}{\ensuremath{\mathcal{D}}\xspace}
\newcommand{\task}{\ensuremath{\mathcal{T}}\xspace}
\newcommand{\eval}{$\textsc{Eval}$\xspace}

A RAG framework typically consists of:
\begin{itemize}[noitemsep, topsep=0pt]
\setlength{\itemsep}{0pt}
    \item A retriever \ir parameterized by $\theta$
    \item A large language model \llm parameterized by $\phi$
    \item A task \task presented as an instruction prompt
    \item A datastore \datastore with a vast number of documents $d$
    \item A user query $q$
    \item The answers $a$ to the query
    \item An evaluation metric \eval determining whether the output generation addresses the query
\end{itemize}

The downstream RAG pipeline generally follows:
\begin{enumerate}[noitemsep, topsep=0pt]
    \setlength{\itemsep}{0pt}
    \item Retrieve the top-$k$ relevant documents from the $\mathcal{D}$ based on $q$, with a relevance function $f_\theta$:
        \begin{align*}
            \{\hat{d}\}_k &= \mathcal{R}_\theta(q, \mathcal{D}, k) \triangleq \underset{d \in \mathcal{D}}{\operatorname{argmax}_k} f_\theta(q, d)
        \end{align*}
    \item Formulate the task-specific prompt $x$ using the query $q$ and the retrieved documents $\{\hat{d}\}_k$:
        \begin{align*}
            x &= \textit{Prompt}_\mathcal{T}(q, \{\hat{d}\}_k)
        \end{align*}
    \item Generate response $\hat{y}$ from input $x$ via LLM:
        \begin{align*}
            \hat{y} &= \mathcal{G}_\phi(x)
        \end{align*}
    \item Evaluate if the generation $\hat{y}$ reflects the answer $a$:
        \begin{align*}
            \textsc{Eval}(\hat{y}) &= \begin{cases}
            1 & \text{if $\hat{y}$ reflects $a$,} \\
            0 & \text{otherwise.}
            \end{cases}
        \end{align*}
\end{enumerate}

\textbf{The Goal of \openrag}: In a RAG system, given an LLM, a datastore, and a task, \openrag\ aims to train the retriever component to maximize the likelihood of generating a response $\hat{y}$ that optimally satisfies the downstream evaluation metric. This can be formulated as:
\begin{gather*} 
\hat{\theta} = \underset{\theta}{\operatorname{argmax}} \; \sum_{\forall q} \textsc{Eval}(\hat{y} \mid \mathcal{R}_\theta, \mathcal{G}_\phi, \mathcal{T}, \mathcal{D}, q)
\end{gather*}
\subsection{Challenges and Prior Work}
\label{sec:prior_work}

\paragraph{Major Challenges.} 
There are two major challenges in training a RAG framework end-to-end via tuning retriever. (i) The primary challenge involves the extreme computational costs associated with deploying such a pipeline in training. These costs mainly arise from two sources: first, the LLMs generate sequences autoregressively, which is inherently resource-intensive; secondly, as $\theta$ updates, the retrieval index need to be rebuilt accordingly, adding further computational demands. (ii) The second challenge is ensuring stable and effective back-propagation of supervision signals from the final outcome of the RAG pipeline to the retriever.

\paragraph{Prior Practices.} 
Prior research~\cite{guu2020retrieval, xu2023retrieval, shi2023replug} has explored the joint training of retrievers with LLMs for RAG. Despite extensive efforts, they often default to learning a universal relevance, where the retrieved document aids in generating the continuation of a natural language input, while neglecting the specific downstream components $\mathcal{T}$, $\mathcal{D}$, $\mathcal{G}_\phi(x)$ and \textsc{Eval}. These general approaches lead to a significant discrepancy as the components used during training do not align with those employed during inference. As a result, these methods often fall short in meeting the specific, nuanced relevance needs of various downstream tasks.

\section{Methodology}

\begin{figure*}[t]
\begin{center}
\includegraphics[width=0.98\textwidth]{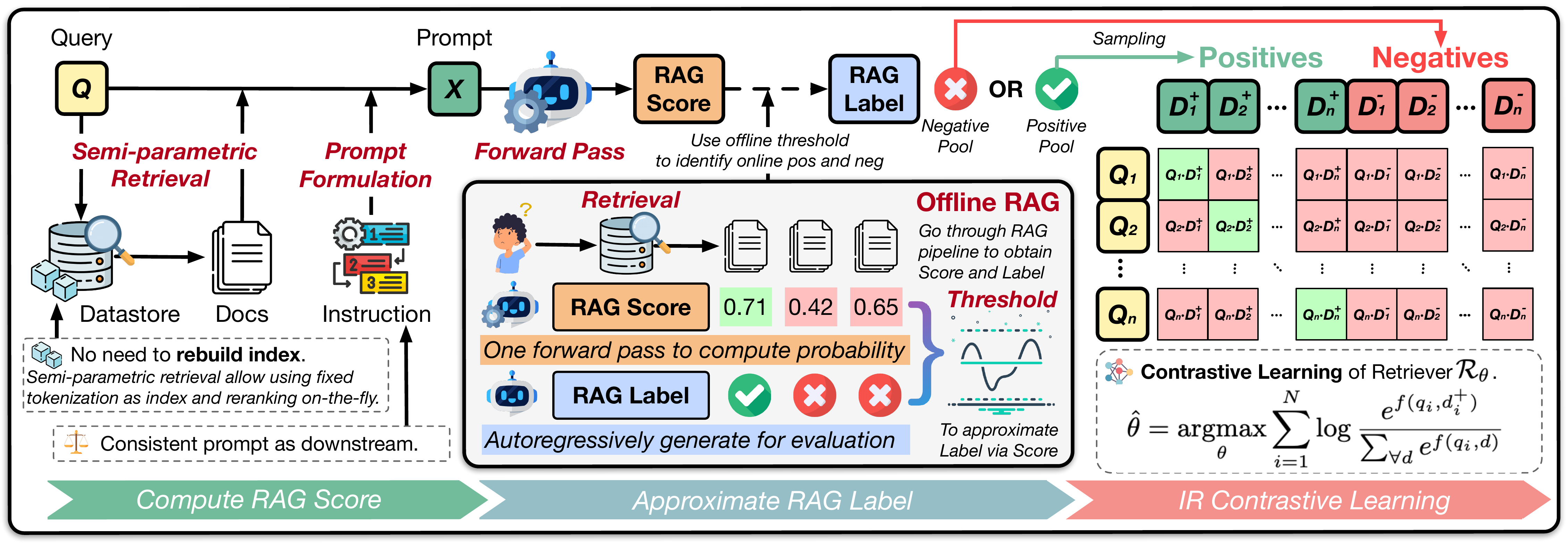}
\vspace{-0.3cm}
\caption{Illustration of the \openrag training process.}
\label{fig:main}
\end{center}
\vspace{-0.5cm}
\end{figure*}
In this section, we introduce \textbf{\openrag}, an \underline{\textbf{OP}}timized \underline{\textbf{EN}}d-to-end \textbf{RAG} framework designed to fine-tune a retriever to capture in-context, \underline{\textbf{open}}-ended relevance, optimizing it for the downstream RAG pipeline.

To summarize, \openrag training comprises two stages: offline RAG and online RAG. The primary goal is to on-the-fly identify positive and negative documents for the contrastive learning of the retriever. An illustration of our framework is depicted in Figure~\ref{fig:main}.

\subsection{Preliminary Concepts}

\textbf{Continuation $y$ and Generation $\hat{y}$.} 
For knowledge-intensive generative tasks, information is aggregated and prompted as input $x$ to a LLM for generation. The expected output could be an answer string $a$ in question-answering tasks or might be a choice label $c$ in reasoning and fact-checking tasks. Here, we refer to the expected output as the ground truth continuation, denoted as $y$, and the actual output generated by the LLM as $\hat{y}$. In a well-performing RAG framework, it is generally expected that $\hat{y} = y$ or that $\hat{y}$ contain or reflect $y$.

\textbf{RAG Label.} 
Given a query $q$, the RAG label $\mathcal{L}_{d}^{q}$ for a document $d$ is a binary value that indicates whether the RAG outcome, when $d$ is used in the context, meets the evaluation metric. The computation involves the following steps:
\begin{gather*}
x = \textit{Prompt}_\mathcal{T}(q, d); \quad \hat{y} = \mathcal{G}_\phi(x) \\
\mathcal{L}_{d}^{q} \triangleq \textsc{Eval}(\hat{y})
\end{gather*}
This assessment is typically based on whether the generated response contains the answers. The computation of RAG labels aligns with downstream inference, which involves autoregressive generation. For a clearer understanding, we provide examples in Appendix~\ref{appendix:raglabels}.

\textbf{RAG Score.} 
Given a query $q$, the RAG score $\mathcal{S}_{d}^{q}$ of a $d$ is the joint probability that LLM generates continuation $y$ with $d$ in context:
\begin{gather*}
x = \textit{Prompt}_\mathcal{T}(q, d) \\
\mathcal{S}_{d}^{q} \triangleq P_{\phi}(y \mid x) = \prod_{\forall t_i \in y} P_{\phi}(t_i \mid t_{<i}, x)
\end{gather*}
Here, $y = (t_1, \ldots, t_n)$ is a sequence of $n$ tokens and $P_{\phi}$ is the function measures the probability of generating the next token or spans. Unlike the RAG label, the computation of the RAG score requires only a single forward pass of the LLM.
\subsection{Offline RAG}

For offline RAG, we follow the traditional RAG pipeline as mentioned in Section~\ref{sec:problem}. Given a query $q$, we retrieve top-$k$ documents and denote this retrieved subset as $\mathcal{D}_q \subset \mathcal{D}$ where $|\mathcal{D}_q|=k$. We then compute the RAG label and score for each retrieved document $d_i$, resulting in the set $\{(q, d_i, \mathcal{L}_{d_i}^{q}, \mathcal{S}_{d_i}^{q}) \}_{i=1}^k$. Based on their RAG labels, $\mathcal{D}_q$ is further divided into a positive pool $\mathcal{D}_q^{+}$ and a negative pool $\mathcal{D}_q^{-}$. In our experiments, we set $k$ to 100 and discard any sample where either pool is empty.

These RAG offline preparation serve two purposes. First, they establish initial positive and negative query-document pairs to warm up the retriever for tasks. Second, they provide insights into the relationship between the RAG score and the RAG label. Specifically, we want to determine when the RAG score is above a certain threshold, the RAG label is 1, and when the RAG score is below a threshold, the label is 0. This relationship will be used to approximate labels via scores during online RAG training, enabling more efficient online construction of positive and negative pairs.

\subsection{Online RAG}
\label{sec:online}
\paragraph{In-training Retrieval.}
During retriever training, as its parameters update, the index needs to be rebuilt accordingly, which incurs significant costs. To address this challenge, we employ the semi-parametric retriever \sidr~\cite{zhou2024semi}. Specifically, \sidr incorporates both a parametric and a non-parametric encoder. The parametric encoder embeds text input $x$ into a sparse representation with $|V|$ dimensions, where each dimension signifies the importance of a token within the language model's vocabulary $V$, denoted as $V_{\theta}(x)$. Conversely, the non-parametric encoder converts $x$ into bag-of-tokens representation, referred to as $V_\text{BoT}(x)$, which is constructed via a tokenizer and is independent of $\theta$. \sidr is strategically trained to allow the embedded query $V_{\theta}(q)$ to search on both an embedding-based index $V_{\theta}(\mathcal{D})$ and a bag-of-tokens index $V_\text{BoT}(\mathcal{D})$.

We adopt the late parametric mechanism of \sidr, which firstly retrieve the top-$m$ documents using the bag-of-tokens index $V_\text{BoT}(\mathcal{D})$, denoted as:
\begin{align*}
    \{\hat{d}\}_m &= \mathcal{R}_\theta(V_\theta(q), V_\text{BoT}(\datastore), m)
\end{align*}
These retrieved documents are then embedded and re-ranked on-the-fly to yield the top-$k$ well-ranked documents, where $k<m$:
\begin{align*}
    \{\hat{d}\}_k &= \mathcal{R}_\theta(V_\theta(q), V_\theta(\{\hat{d}\}_m), k)
\end{align*}
In this case, our in-training retrieval does not require index updates, and the relevance is based on the up-to-date parameters. For late parametric mechanism, we set $m=k=20$ to reduce training cost. More details of \sidr can be found in Appendix~\ref{appendix:sidr}.

\paragraph{Identifying Positives and Negatives On-the-fly.}
During training, we denote the pool of top-$k$ retrieved documents as $\hat{\mathcal{D}}_q$. Our goal is to divide $\hat{\mathcal{D}}_q$ into a positive pool $\hat{\mathcal{D}}_q^{+}$ and a negative pool $\hat{\mathcal{D}}_q^{-}$ without the need for autoregressive generation. We present how to achieve this identification in two generation scenarios.

For \underline{\textit{free-form generation}}, such as in question answering tasks, the continuation $y$ typically consists of a multi-token answer string. We identify a retrieved document $\hat{d}$ as positive if its RAG score surpasses the highest RAG score in the offline negative pool $\mathcal{D}_q^{-}$ and as negative if it is below the lowest RAG score in the offline positive pool $\mathcal{D}_q^{+}$; otherwise, it is excluded: 
\begin{align*}
    \mathcal{\hat{L}}_{\hat{d}}^{q} &= \begin{cases}
    1, & \text{if } \mathcal{S}_{\hat{d}}^{q} > \max \{\mathcal{S}_{d}^{q} \mid \forall d \in \mathcal{D}_q^{-} \} \\
    0, & \text{if } \mathcal{S}_{\hat{d}}^{q} < \min \{\mathcal{S}_{d}^{q} \mid \forall d \in \mathcal{D}_q^{+} \} \\
    \text{None}, & \text{otherwise}
    \end{cases}
\end{align*}
Here, we use $\mathcal{\hat{L}}$ to denote the online RAG label, as it involves certain approximation. The approximation is based on the \emph{assumption} that a higher RAG score correlates with an increased probability that the generated output $\hat{y}$ will match or reflect the target $y$. This strategy aims to reduce computational costs, enabling low-resource institutions and individuals to conduct retriever training. If computational resources are not a limitation, ideally, one could perform autoregressive generation and evaluation on-the-fly or employ a larger LLM for identification purposes. We provide further discussion and verification of this assumption in Appendix~\ref{appendix:ragscores}.

For \underline{\textit{closed-set generation}}, such as in multiple-choice reasoning or fact-checking tasks, the continuation $y$ is typically a single-token choice label or can be prompted as such. In this case, we can relax the assumptions:
\begin{align*}
    \mathcal{\hat{L}}_{\hat{d}}^{q} &= \begin{cases}
    1, 
    & \text{if } P_{\phi}(c_i \mid x) > \max \{P_{\phi}(c_j \mid x) \mid \forall j \neq i \} \\
    0, 
    & \text{otherwise.} 
    \end{cases}
\end{align*}
Here, $x$ is the input prompt and $c_i$ is the correct single-token choice while $c_j$ are the incorrect choices. This setup checks whether LLM is more likely to generate $c_i$ as the next token following $x$ instead of $c_j$, when $\hat{d}$ is used in context. 

For both scenarios, if a query has multiple correct continuation $y$ (answers or choices), each $y$ is treated as an individual entry. If $\hat{d}$ succeeds on at least one of these entries, we label it as positive; if it fails all of them, we label it as negative.

\paragraph{Sampling and Cache.} 
During the online phase, we retrieve the top-$k$ documents and compute their RAG scores to approximate RAG labels, processing them in descending order of retrieval relevance. We stop this process at the first document classified as negative. We then use this highest relevant negative, denoted as $\hat{d}^{-}$, and randomly select one positive $\hat{d}^{+}$ from the pool $\mathcal{\hat{D}}_q^{+}$. If either is unavailable, we fallback to random sampling from offline positive pool $\mathcal{D}_q^{+}$ or negative $\mathcal{D}_q^{-}$. To avoid redundant calculations, we cache all the online scores and labels $\{(q, \hat{d}_i, \mathcal{\hat{L}}_{\hat{d}_i}^{q}, \mathcal{\hat{S}}_{\hat{d}_i}^{q}) \}$ for reuse.

%\subsection{RAG-supervised IR via Contrastive Learning}
\subsection{Contrastive Learning}

Throughout our offline and online efforts, our objective is to acquire high-quality positive and negative query-document pairs for the contrastive learning~\cite{jaiswal2020survey} of the retriever $\mathcal{R}_\theta$. Positives and negatives are determined by their impact on the RAG output; specifically, their ability to enable the RAG framework to generate the correct continuation that meets the criteria of the evaluation metric. This ensures that supervision signals are propagated from the end of the RAG pipeline back to the retriever.

Our training objective remains the same as \sidr to maintain its ability for late parametric. Given a batch $B$ that consist of $N$ samples, each sample consists of a query $q_i$, a positive document $d_i^{+}$, and a negative document $d_i^{-}$. Our training objective aims to maximize the similarity of positive query-document pairs $f(q_i, d_i^{+})$ for all instances $i$, while minimize the similarity of all negative pairs, denoted as $f(q_i, d)$ for all $d \neq d_i^{+}$. The contrastive loss can be defined as follows:
\begin{small}
\begin{equation}
\begin{split}
L(q, d) = &-\sum_{i=1}^{N}(
\log{
\underbrace{
    \frac{e^{{f(q_i,d_i^{+})}}}
        {\sum_{\forall d \in B} e^{f(q_i,d)}}}
_\text{q-to-d}}
+ 
\log{
\underbrace{
    \frac{e^{f(d_i^{+},q_i)}}{\sum_{\forall q \in B} e^{f(d_i^{+},q_i)}}}
_\text{d-to-q}}) \nonumber
\end{split}
\end{equation}
\end{small}
The final loss integrates contrastive loss of both parametric and semi-parametric components:
\begin{small}
\begin{equation}
\begin{split}
L_{\text{para}}(q,d) &= L(V_{\theta}(q), V_\theta(d))
\\
L_{\text{semi-para}}(q,d) &= L(V_\theta(q), V_{\text{BoT}}(d))/2 + L(V_{\text{BoT}}(q), V_\theta(d))/2
\\
L_{\text{final}}(q,d) &= L_{\text{para}}(q,d) + L_{\text{semi-para}}(q,d)
\nonumber
\end{split}
\end{equation}
\end{small}

\iffalse
\begin{small}
\begin{equation}
\begin{split}
\hat{\theta} = \operatorname*{argmax}_\theta & \sum_{i=1}^{N}\log{\frac{e^{{f(q_i,d_i^{+})}}}{\sum_{\forall d} e^{f(q_i,d)}}} \nonumber
\end{split}
\end{equation}
\end{small}

$\mathcal{R}_\theta$
\fi

\section{Experiments}
\begin{table*}[ht]
  \centering
  \caption{Main results of \openrag and other RAG baselines on 4 datasets, using top-1 and top-10 retrieved documents in context. \textbf{Bold}: best RAG method that does not involve LLM tuning. $\Delta$: \textcolor{mygreen}{improvement} or \textcolor{myred}{decline}; $\blacktriangle$: baseline that below methods compare with; $\dagger$:~reproduction from other works; $\ddagger$:~our reproduction; $\S$:~has accessed the training split of the dataset.}
\small
\scalebox{0.78}{
\setlength{\tabcolsep}{0.8mm}{
\begin{tabular}{@{}lc>{\columncolor{gray!20}}cc>{\columncolor{gray!20}}c|c>{\columncolor{gray!20}}cc>{\columncolor{gray!20}}c|c>{\columncolor{gray!20}}cc>{\columncolor{gray!20}}c|c>{\columncolor{gray!20}}cc>{\columncolor{gray!20}}c|@{}}
\toprule
\textbf{Task Type ($\rightarrow$)} & \multicolumn{8}{c}{\textbf{Free-form}} & \multicolumn{8}{c}{\textbf{Closed-set}} \\
\cmidrule(lr){2-9}\cmidrule(lr){10-17}
\textbf{\quad Dataset ($\rightarrow$)} & \multicolumn{4}{c}{\textbf{NQ}} & \multicolumn{4}{c}{\textbf{TriviaQA}} & \multicolumn{4}{c}{\textbf{PubHealth}} & \multicolumn{4}{c}{\textbf{ARC-C}} \\
\cmidrule(lr){2-5}\cmidrule(lr){6-9}\cmidrule(lr){10-13}\cmidrule(lr){14-17}
\textbf{Method ($\downarrow$) \quad Metrics ($\rightarrow$)} & \textbf{1-doc} & $\Delta$ & \textbf{10-doc} & $\Delta$ & \textbf{1-doc} & $\Delta$ & \textbf{10-doc} & $\Delta$ & \textbf{1-doc} & $\Delta$ & \textbf{10-doc} & $\Delta$ & \textbf{1-doc} & $\Delta$ & \textbf{10-doc} & $\Delta$ \\
\midrule
\multicolumn{17}{c}{\textit{\textbf{Standard RAG}}} \\
\midrule
\multicolumn{17}{l}{\textit{\textbf{Baseline IR}}} \\
\quad Llama3$_\text{8B}$ + \sidrms & 34.4 & $\blacktriangle$ & 37.6 & $\blacktriangle$ & 62.0 & $\blacktriangle$ & 62.5 & $\blacktriangle$ & 63.5 & $\blacktriangle$ & 64.9 & $\blacktriangle$ & 56.9 & $\blacktriangle$ & 57.5 & $\blacktriangle$ \\
\quad Llama3$_\text{8B}$ + \sidrnq & $\S$42.7 & \increase{8.3} & $\S$41.6 & \increase{4.0} & -- & -- & -- & -- & -- & -- & -- & -- & -- & -- & -- & -- \\
\multicolumn{17}{l}{\textit{\textbf{Advanced IR}}} \\
\quad Llama3$_\text{8B}$ + \contrieverms & 36.5 & \increase{2.1} & 38.3 & \increase{0.7} & 60.7 & \decrease{1.3} & 60.6 & \decrease{1.9} & 63.1 & \decrease{0.4} & 62.9 & \decrease{2.0}  & \textbf{58.1} & \textbf{\increase{1.2}} & \textbf{58.9} & \textbf{\increase{1.4}} \\
\quad Llama3$_\text{8B}$ + \efive & \textbf{$\S$43.2} & \textbf{\increase{8.8}} & \textbf{$\S$41.8} & \textbf{\increase{4.2}} & 63.2 & \increase{1.2} & 61.4 & \decrease{1.1} & 64.7 & \increase{1.2} & 63.7 & \decrease{1.2} & 58.0 & \increase{1.1} & 58.1 & \increase{0.6} \\
\midrule
\multicolumn{17}{c}{\textit{\textbf{RAG with IR tuning}}} \\
\midrule
\quad $\dagger\textsc{RePlug}_\text{Llama2-7B}$ (3-doc, \citet{yue2024synergistic}) & -- & -- & -- & -- & -- & -- & -- & -- & -- & -- & 41.7 & -- & -- & -- & 47.2 & -- \\
\multicolumn{17}{l}{\textit{\textbf{Ours}}} \\
\quad \openrag (\sidrms) & 39.8 & \increase{5.4} & 40.9 & \increase{3.3} & \textbf{65.8} & \textbf{\increase{3.8}} & \textbf{66.2} & \textbf{ \increase{3.7}} & \textbf{69.5} & \textbf{\increase{6.0}} & \textbf{69.3} & \textbf{\increase{4.4}} & \textbf{58.1} & \textbf{\increase{1.2}} & 58.3 & \increase{0.8} \\
\quad \openrag (\sidrnq) & \textbf{$\S$44.1} & \textbf{\increase{9.7}} & \textbf{$\S$44.7} & \textbf{\increase{7.1}} & -- & -- & -- & -- & -- & -- & -- & -- & -- & -- & -- & -- \\
\midrule
\multicolumn{17}{c}{\textit{\textbf{RAG with LLM tuning}}} \\
\midrule
\quad Llama3-Instruct$_\text{8B}$ +  \sidrms & 41.2 & \increase{6.8} & 52.1 & \increase{14.5} & 65.2 &  \increase{3.2} & 73.3 & \increase{10.8} & 67.2 & \increase{3.7} & 71.8 & \increase{6.9} & 72.1 & \increase{15.2} & 75.5 & \increase{18.0} \\
\quad $\textsc{Self-RAG}_\text{Llama2-7B}$~\citep{asai2023self} & -- & -- & -- & -- & -- & -- & 66.4 & \increase{3.9} & -- & -- & 72.4 & \increase{7.5} & -- & -- & 67.3 & \increase{9.8} \\
\quad $\dagger\textsc{Self-RAG}_\text{Mistral-7B}$~\citep{wang2024speculative} & -- & -- & -- & -- & -- & -- & 64.8 & \increase{2.3} & -- & -- & 72.4 & \increase{7.5} & -- & -- & 74.9 & \increase{17.4} \\
\quad $\dagger\textsc{Self-RAG}_\text{Llama3-8B}$~\citep{zhang2024raglab} & -- & -- & -- & -- & -- & -- & 56.4 & \decrease{6.1} & -- & -- & 67.8 & \increase{2.9} & -- & -- & 58.0 & \increase{0.5} \\
\quad $\ddagger\textsc{Self-RAG}_\text{Llama3-8B}$ + \sidrms & 30.8 & \decrease{3.6} & 37.0 & \decrease{0.6} & 51.0 & \decrease{11.0} & 57.7 & \decrease{4.8} & 64.2 & \increase{0.7} & 64.0 & \decrease{0.9} & 58.9 & \increase{2.0} & 59.1 & \increase{1.6} \\
\midrule
\multicolumn{17}{c}{\textit{\textbf{Transferring \openrag to other LLM}}} \\
\midrule
\quad Llama3-Instruct$_\text{8B}$ +  \sidrms & 41.2 & $\blacktriangle$ & 52.1 & $\blacktriangle$ & 65.2 & $\blacktriangle$ & 73.3 & $\blacktriangle$ & 67.2 & $\blacktriangle$ & 71.8 & $\blacktriangle$ & 72.1 & $\blacktriangle$ & 75.5 & $\blacktriangle$ \\
\quad Llama3-Instruct$_\text{8B}$ +  \openrag(\sidrms) & 43.6 & \increase{2.4} & 54.7 & \increase{2.6} & 65.6 &  \increase{0.4} & 73.8 &  \increase{0.5} & 65.2 & \decrease{2.0} & 66.1 & \decrease{5.7} & 71.9 & \decrease{0.2} & 75.0 & \decrease{0.5} \\
\quad Phi-3-mini-4k-instruct$_\text{3.8B}$ + \sidrms & 40.6 & $\blacktriangle$ & 49.2 & $\blacktriangle$ & 64.6 & $\blacktriangle$ & 69.2 & $\blacktriangle$ & 48.2 & $\blacktriangle$ & 57.6 & $\blacktriangle$ & 84.9 & $\blacktriangle$ & 84.3 & $\blacktriangle$ \\
\quad Phi-3-mini-4k-instruct$_\text{3.8B}$ +  \openrag(\sidrms) & 43.4 & \increase{2.8} & 50.3 & \increase{1.1} & 65.6 &  \increase{1.0} & 70.4 &  \increase{1.2} & 45.3 & \decrease{2.9} & 54.4 & \decrease{3.2} & 85.1 & \increase{0.2} & 84.6 & \increase{0.3} \\
\quad Mistral-Instruct$_\text{7B}$ + \sidrms & 37.5 & $\blacktriangle$ & 48.0 & $\blacktriangle$ & 58.2 & $\blacktriangle$ & 57.1 & $\blacktriangle$ & 50.1 & $\blacktriangle$ & 57.4 & $\blacktriangle$ & 69.7 & $\blacktriangle$ & 71.5 & $\blacktriangle$ \\
\quad Mistral-Instruct$_\text{7B}$ +  \openrag(\sidrms) & 40.5 & \increase{3.0} & 49.4 & \increase{1.4} & 59.8 &  \increase{1.6} & 57.6 &  \increase{0.5} & 46.7 & \decrease{3.4} & 54.6 & \decrease{2.8} & 69.2 & \decrease{0.5} & 70.6 & \decrease{0.9} \\
\bottomrule
\end{tabular}%
}}
\label{tab:main}%
\vspace{-0.28cm}
\end{table*}

\subsection{Experimental Setup}
\label{sec:setup}
\textbf{Tasks and Datasets.} 
We evaluate \openrag on four public RAG benchmarks. For free-form generation, we utilize Natural Questions (NQ; \citeauthor{kwiatkowski2019natural}, \citeyear{kwiatkowski2019natural}) and TriviaQA (TQA; \citeauthor{joshi2017triviaqa}, \citeyear{joshi2017triviaqa}), two well-established open-domain QA datasets. For closed-set generation, we employ the PubHealth~\cite{kotonya2020explainable} dataset for fact-checking tasks, and the ARC-Challenge~\cite{clark2018think} dataset for multiple-choice reasoning. More information about the datasets can be found in Appendix~\ref{appendix:dataset}. 

We exclude long-form generation datasets as we use the probability of continuation to approximate RAG performance, which may not align well with such tasks. Additionally, certain datasets, such as PopQA~\cite{mallen2023not}, which only offer a test split, are also excluded.

\textbf{Evaluation Metrics.}
Following previous works~\cite{asai2023self, mallen2023not}, we use accuracy as the evaluation metric and report results on the test set. In IR scenarios, accuracy is measured by whether the retrieved documents contain the expected answers, while in RAG scenarios, it is assessed based on the generated output. Since our training uses 1 document in context while existing research generally uses 10 for RAG, we report accuracy with both 1 and 10 documents in context for comparison.

\textbf{Implementation Details.} Our RAG system employs the LLM Llama3-8b~\cite{dubey2024llama} with the retriever \sidrms~\cite{zhou2024semi} that trained on MS MARCO dataset~\cite{bajaj2016ms}. We use the same English Wikipedia datastore and prompt as those open-sourced by \selfrag, detailed in Appendix~\ref{appendix:prompt}. During training, we train the retriever for each dataset for 80 epochs, aligning with the training duration used for \sidrms. We use a batch size of 128 and an AdamW optimizer~\citep{loshchilov2018decoupled} with a learning rate of $2 \times 10^{-5}$. The training process is divided into two phases: the first half involves a warm-up phase using offline positives and negatives, while the second half transitions to in-training retrieval, primarily using the positives and negatives identified on-the-fly. During inference, we set the maximum number of generated token to be 100 for free-form generation while 20 for closed-set generation.

\textbf{Training Costs.}
Our experiments are conducted with 4 NVIDIA A100 GPUs. Both offline RAG preparation and online RAG training take less than one day, depending on the number of queries in the datasets. We leverage vLLM~\cite{kwon2023efficient} to accelerate offline generation.

\textbf{Baselines.}
We consider the baselines detailed below, with additional model information provided in Appendix~\ref{appendix:models}. 
(1)~\underline{\emph{Standard RAG with advanced IR}}: RAG frameworks using Llama3-8b and state-of-the-art retrievers \efive~\cite{wang2022text} and \contrieverms~\cite{izacard2021towards}. We refer to \openrag(\sidrnq) and \openrag(\sidrms) as our framework utilizing \sidrnq and \sidrms as the initial retriever, respectively. Unless explicitly stated otherwise, \openrag refers to \openrag(\sidrms). For a fair comparison, we compare \efive with \openrag(\sidrnq), both of which have access to the query-document pairs from the NQ training split.
(2)~\underline{\emph{RAG with IR tuning}}: RAG frameworks that incorporate a tunable IR component. We compare against \replug~\cite{shi2023replug}, which uses part of a sequence as query to retrieve documents which maximize the generation likelihood of the remaining part. Since the model weights are not publicly available, we reference a reproduction by \cite{yue2024synergistic} that uses the top-3 retrieved documents in context. 
(3)~\underline{\emph{RAG with LLM tuning}}: RAG frameworks that incorporate RAG-oriented or instruction-tuned LLMs, which typically require more resources for tuning an 8B LLM. We compare with \selfrag~\cite{asai2023self} using Llama2-7B, along with some reproductions~\cite{zhang2024raglab,wang2024speculative} employing more recent LLMs. Our primary comparison with \selfrag and its variants is designed to ensure a controlled and fair evaluation, as we adhere to the same prompts and downstream evaluation pipeline.
(4)~\underline{\emph{Transferring to other LLMs}}: We compare the RAG framework using different LLMs, such as Llama3-Instruct$_\text{8B}$~\cite{dubey2024llama}, Phi-3-mini-4k-instruct$_\text{3.8B}$~\cite{abdin2024phi}, Mistral-Instruct$_\text{7B}$~\cite{jiang2023mistral}, along with \sidrms before and after tuning. This setup is designed to evaluate whether the learned in-context relevance transfers across different LLMs.

\subsection{Main Experiments}
Table~\ref{tab:main} presents results of \openrag and other baselines. The key findings are summarized as follows:

\textbf{End-to-end tuning effectively improves the retriever in RAG scenarios, surpassing existing SOTA retrievers.} 
Unlike \efive and \contrieverms, which require both extensive pre-training and human-labeled query-document pairs, \openrag improves the initial retriever using only downstream queries, achieving better automation and training efficiency. Our approach leads to a notable 4.0\% enhancement in performance beyond the original \sidrms and consistently achieves a 2.1\% better outcome than the SOTA retrievers. For PubHealth, the improvement reaches up to 6\%, a significant value that even using instruction-tuned LLMs cannot achieve. For ARC, the modest improvement can be attributed to its limited number of training samples, only a few hundred, compared to other datasets containing tens of thousands. These results demonstrate that, despite approximation, the learned in-context relevance is more effective than the inconsistent relevance derived from existing datasets. In Appendix~\ref{appendix:gap}, we show that improve the retriever for RAG scenarios may degrade its performance in traditional IR scenarios, further reinforcing this inconsistency.

\textbf{Relevance learning constitutes a valuable yet overlooked dimension for improving the RAG system.} Reproductions of \selfrag using Llama3-8B by other works~\cite{zhang2024raglab,wang2024speculative} and ourselves have not yielded consistent improvements. This suggests that despite the substantial training expenses, enhancing RAG through tuning LLM requires extensive customization and does not reliably generalize. In contrast, tuning a smaller-sized retriever can lead to comparable, or in some cases, superior improvements over those achieved by RAG-oriented or instruction-tuned 8B LLMs on specific datasets. Importantly, learning an in-context retriever does not conflict with LLM enhancements, offering a complementary avenue for improving the RAG system.

\textbf{The learned in-context retriever can be transferred to other LLMs for free-form generation tasks.} Our results show that \openrag, initially co-trained with Llama3-8b, enhances other LLMs such as Llama3-Instruct-8B, Phi-3-mini-4k-instruct, and Mistral-Instruct in free-form generation tasks. However, for closed-set generation tasks, this transferability does not consistently hold. Despite the limitations, \openrag significantly enhances performance of PubHealth by a large margin. We hypothesize that closed-set tasks, where the continuation is a single token, are easier to optimize due to less approximation involved. Consequently, the retriever learns a very specific relevance tailored to the particular LLM prediction of the next token, complicating its transferability. Therefore, we recommend end-to-end tuning on a LLM-by-LLM basis to potentially improve outcomes for these tasks.

\subsection{Ablation Study}
Compared to prior works, our main differences include (i) employing contrastive learning instead of KL divergence to induce supervision signals from the LLM to the IR, and (ii) using late parametric to avoid periodic re-indexing. We systematically analyze these factors in this section.

\begin{figure}[th]
\begin{center}
\includegraphics[width=0.9\columnwidth]{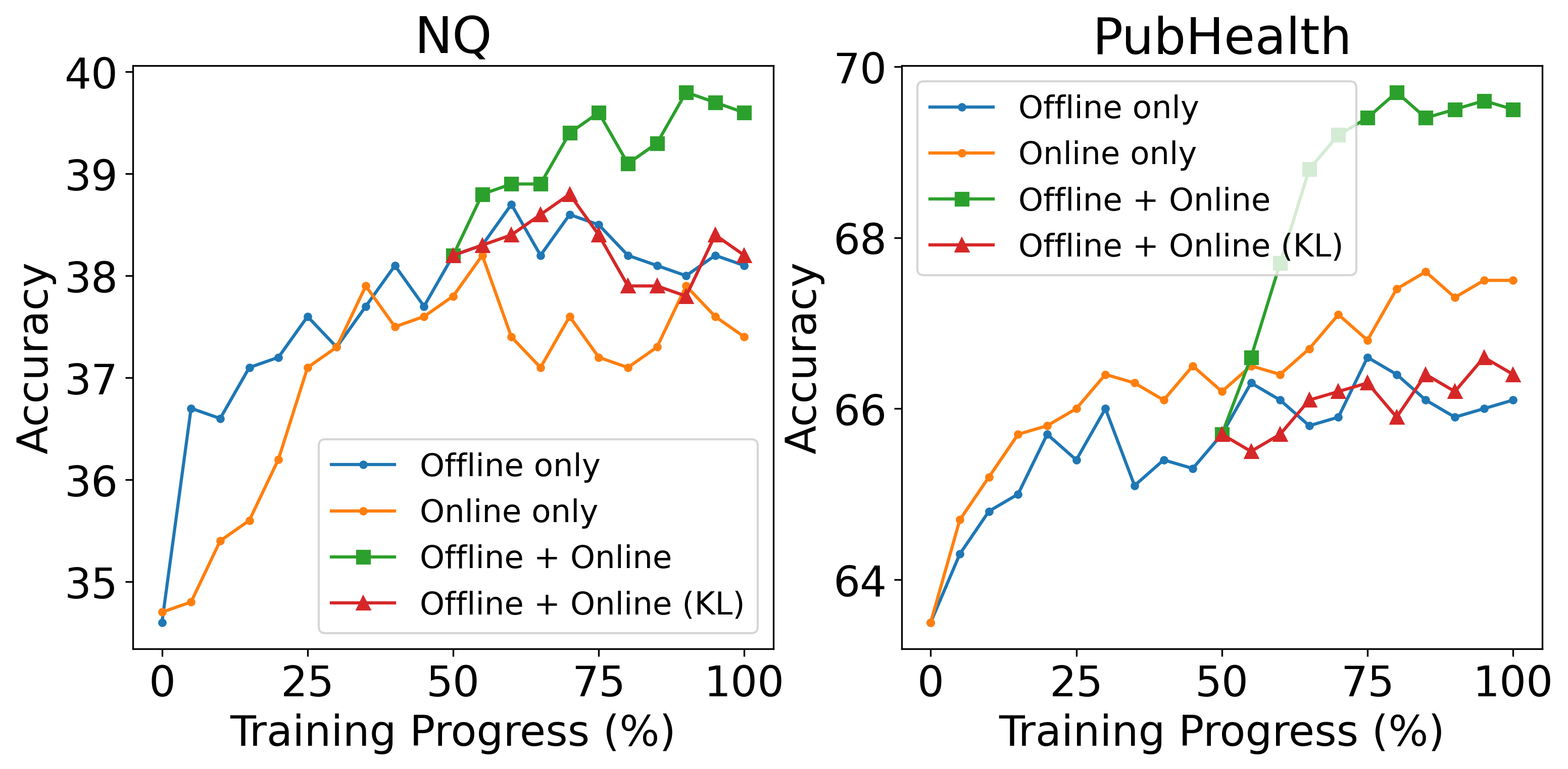}
\vspace{-0.1cm}
\caption{Ablation studies on NQ and Pubhealth datasets.}
\label{fig:abl}
\end{center}
\end{figure}
\vspace{-0.3cm}

As shown in Figure~\ref{fig:abl}, we conducted an ablation study on NQ and PubHealth with several setup: our method is labeled as \emph{[offline+online]}, where \emph{[offline-only]} represents using only the offline positives and negatives for contrastive learning, and \emph{[online-only]} indicates that we do not use any warmup. We also explore using KL divergence \emph{[offline+online(KL)]} instead of contrastive learning.

\textbf{Offline versus Online.} 
During the warmup stage, documents are retrieved using the initial parameters $\theta$. During the in-training retrieval stage, they are retrieved using the up-to-date parameters $\theta'$. We assess the improvements provided by the in-training retrieval stage. As shown in Figure~\ref{fig:abl}, relying solely on either \emph{[offline-only]} or \emph{[online-only]} can lead to suboptimal improvements, proving to be less effective than a combination of a warmup phase followed by online in-training retrieval \emph{[offline+online]}. This observation echoes the conclusions of prior research~\cite{zhou2024semi}, which indicates that warming up the retriever to initially capture the in-task relevance, followed by in-training retrieval to continuously explore potential positives and challenging negatives in the datastore, can significantly enhance performance.

\textbf{Contrastive Learning versus KL-Divergence.} 
Prior works~\cite{shi2023replug, guu2020retrieval} have employed KL divergence to align query-document relevance with the distribution of generation likelihood. Our experiments indicate that while KL divergence leads to improvements, these benefits quickly stabilize and the overall enhancement falls short of our method. Unlike our approach, which employs contrastive learning requiring efforts to identify positives and negatives, KL divergence alignment offers a straightforward but potentially overly restrictive solution. On one hand, in RAG scenarios, documents are delivered to LLMs, differing from IR scenarios where documents must be well-ranked before being presented to users. For a proficient LLM, including even a single useful document in the context window should suffice~\cite{cuconasu2024power}.
On the other hand, similar works in knowledge distillation~\cite{gou2021knowledge}, which uses cross-encoder scores to guide bi-encoder training, demonstrate that improvements for bi-encoders are limited and cannot match the performance of cross-encoder rerankers. Consequently, the prevalent industry practice of retrieve-then-rerank~\cite{gupta2018retrieve} underscores the current limitations of retrievers in capturing complex relationships. We believe that the distribution of generation likelihood from LLMs is too complex for these small-sized retriever to accurately capture, thereby resulting in less improvement.

\textbf{Late Parametric versus Periodic Re-indexing.}
Due to page limitations, we detail our comparison of different in-training retrieval methods in Appendix~\ref{appendix:latepara}. This comparison particularly focuses on the late parametric method versus prior solutions that utilize an embedding index and require periodic re-indexing. Our results indicate that the late parametric method not only leads to better improvements but also reduces training costs and simplifies the implementation. We believe that the high costs and complex implementation associated with periodic re-indexing have prevented previous research from effectively training retrievers on a task-by-task basis, using consistent instructions, LLMs, and datastores tailored to downstream tasks, ultimately leading to less effective results.

\subsection{Cost-Effectiveness Analysis}
Regarding training costs, the primary expense comes from computing the RAG scores using the LLM. In Table~\ref{tab:cost}, we report the number of documents required to compute RAG scores on-the-fly during training.
\begin{table}[ht]
\centering
\scalebox{0.8}{
\begin{tabular}{ccccc}
\hline
 & \textbf{NQ} & \textbf{TriviaQA} & \textbf{PubHealth} & \textbf{ARC} \\ \hline
\textbf{nDoc} & 20 & 18 & 128 & 15 \\
\textbf{Improv.}   & +5.4\% & +3.8\% & +6.0\% & +1.2\% \\ \hline
\end{tabular}
}
\caption{Number of documents required on-the-fly RAG score computation and the improvement for each task.}
\label{tab:cost}
\vspace{-0.2cm}
\end{table}
\vspace{-0.2cm}

Throughout training, each query encounters between 15 to 128 unscored documents, depending on the task, requiring LLM forward passes to compute RAG scores on-the-fly. This process incurs a manageable cost, typically amounting to hours rather than days. We also observe a positive correlation between the number of documents processed and the performance improvements of \openrag. Notably, the PubHealth dataset requires more documents to compute the RAG score online, resulting in the most significant improvement. This suggests that encountering more unscored documents indicates a larger gap in relevance between the initial and the learned retriever, highlighting the presence of more potentially useful documents in the datastore that could be leveraged by in-context retrieval learning.

\section{Related Works}
\textbf{Retrieval-augmented Generation (RAG).}
The RAG system combines LLMs, retrievers, and datastores, each contributing to performance improvement. Significant research has focused on improving RAG by tuning LLMs to address challenges such as enhancing on-demand retrieval~\cite{asai2023self, jeong2024adaptive}, optimizing response efficiency~\cite{wang2024speculative}, and enabling self-reasoning capabilities~\cite{li2024alr}. Additional efforts have explored building domain-specific~\cite{wang2024coderag} or large datastores~\cite{shao2024scaling}. 
While some studies focus on retrieval, exploring adaptive retrieval strategies~\cite{wang2024astute, wang2024searching} and leveraging LLMs to develop stronger retrievers~\cite{guu2020retrieval, shi2023replug}, research on end-to-end relevance learning for RAG scenarios remains limited. Our work addresses this gap, paving the way for new advancements in RAG systems.

\textbf{Relevance Learning.}
Relevance learning is an important and long-established area of research. Traditionally, text relevance has been measured by heuristic rules based on term overlap, as seen in the widely-used BM25~\cite{robertson2009probabilistic}. With advances in deep learning, neural retrievers have emerged~\cite{karpukhin2020dense}, learning relevance from human-annotated datasets~\cite{kwiatkowski2019natural}. Further research has explored pre-training retrievers using weakly supervised text pairs, such as cropped text spans within documents~\cite{izacard2021towards} and relational text pairs extracted from web data~\cite{zhou2022hyperlink, wang2022text}, to enable retrievers to learn general relevance. This general relevance can then be refined to task-specific and domain-specific relevance through downstream fine-tuning, resulting in improved performance. Our method falls within these advancements, where the LLM acts as a container of general relevance, providing on-the-fly supervision of specific in-context relevance for relevance learning.

\section{Conclusion}
In this work, we show that traditional retrieval relevance derived from QA datasets can be inconsistent in RAG scenarios. To bridge this gap, we introduce \openrag, a RAG framework that learns in-context retrieval end-to-end for downstream tasks. Our framework consistently outperforms RAG frameworks using SOTA retrievers and several that tune an 8B LLM. This highlights the significant potential of retrieval learning to improve RAG performance.

\iffalse
\section*{Impact Statement}
This paper presents work whose goal is to advance the field of 
Machine Learning. There are many potential societal consequences 
of our work, none which we feel must be specifically highlighted here.
\fi

\bibliography{example_paper}

\begin{thebibliography}{59}
\providecommand{\natexlab}[1]{#1}
\providecommand{\url}[1]{\texttt{#1}}
\expandafter\ifx\csname urlstyle\endcsname\relax
  \providecommand{\doi}[1]{doi: #1}\else
  \providecommand{\doi}{doi: \begingroup \urlstyle{rm}\Url}\fi

\bibitem[Abdin et~al.(2024)Abdin, Aneja, Awadalla, Awadallah, Awan, Bach, Bahree, Bakhtiari, Bao, Behl, et~al.]{abdin2024phi}
Abdin, M., Aneja, J., Awadalla, H., Awadallah, A., Awan, A.~A., Bach, N., Bahree, A., Bakhtiari, A., Bao, J., Behl, H., et~al.
\newblock Phi-3 technical report: A highly capable language model locally on your phone.
\newblock \emph{arXiv preprint arXiv:2404.14219}, 2024.

\bibitem[Arora et~al.(2023)Arora, Lewis, Fan, Kahn, and R{\'e}]{arora2023knowledge}
Arora, S., Lewis, P., Fan, A., Kahn, J., and R{\'e}, C.
\newblock Reasoning over public and private data in retrieval-based systems.
\newblock \emph{Transactions of the Association for Computational Linguistics}, 11:\penalty0 902--921, 2023.

\bibitem[Asai et~al.(2023)Asai, Wu, Wang, Sil, and Hajishirzi]{asai2023self}
Asai, A., Wu, Z., Wang, Y., Sil, A., and Hajishirzi, H.
\newblock Self-rag: Learning to retrieve, generate, and critique through self-reflection.
\newblock \emph{arXiv preprint arXiv:2310.11511}, 2023.

\bibitem[Ayyamperumal \& Ge(2024)Ayyamperumal and Ge]{ayyamperumal2024current}
Ayyamperumal, S.~G. and Ge, L.
\newblock Current state of llm risks and ai guardrails.
\newblock \emph{arXiv preprint arXiv:2406.12934}, 2024.

\bibitem[Bajaj et~al.(2016)Bajaj, Campos, Craswell, Deng, Gao, Liu, Majumder, McNamara, Mitra, Nguyen, et~al.]{bajaj2016ms}
Bajaj, P., Campos, D., Craswell, N., Deng, L., Gao, J., Liu, X., Majumder, R., McNamara, A., Mitra, B., Nguyen, T., et~al.
\newblock Ms marco: A human generated machine reading comprehension dataset.
\newblock \emph{arXiv preprint arXiv:1611.09268}, 2016.

\bibitem[Chan et~al.(2024)Chan, Xu, Yuan, Luo, Xue, Guo, and Fu]{chan2024rq}
Chan, C.-M., Xu, C., Yuan, R., Luo, H., Xue, W., Guo, Y., and Fu, J.
\newblock Rq-rag: Learning to refine queries for retrieval augmented generation.
\newblock \emph{arXiv preprint arXiv:2404.00610}, 2024.

\bibitem[Clark et~al.(2018)Clark, Cowhey, Etzioni, Khot, Sabharwal, Schoenick, and Tafjord]{clark2018think}
Clark, P., Cowhey, I., Etzioni, O., Khot, T., Sabharwal, A., Schoenick, C., and Tafjord, O.
\newblock Think you have solved question answering? try arc, the ai2 reasoning challenge.
\newblock \emph{arXiv preprint arXiv:1803.05457}, 2018.

\bibitem[Cuconasu et~al.(2024{\natexlab{a}})Cuconasu, Trappolini, Siciliano, Filice, Campagnano, Maarek, Tonellotto, and Silvestri]{cuconasu2024power}
Cuconasu, F., Trappolini, G., Siciliano, F., Filice, S., Campagnano, C., Maarek, Y., Tonellotto, N., and Silvestri, F.
\newblock The power of noise: Redefining retrieval for rag systems.
\newblock In \emph{Proceedings of the 47th International ACM SIGIR Conference on Research and Development in Information Retrieval}, pp.\  719--729, 2024{\natexlab{a}}.

\bibitem[Cuconasu et~al.(2024{\natexlab{b}})Cuconasu, Trappolini, Siciliano, Filice, Campagnano, Maarek, Tonellotto, Silvestri, et~al.]{cuconasu2024rethinking}
Cuconasu, F., Trappolini, G., Siciliano, F., Filice, S., Campagnano, C., Maarek, Y., Tonellotto, N., Silvestri, F., et~al.
\newblock Rethinking relevance: How noise and distractors impact retrieval-augmented generation.
\newblock In \emph{CEUR WORKSHOP PROCEEDINGS}, volume 3802, pp.\  95--98. CEUR-WS, 2024{\natexlab{b}}.

\bibitem[Devlin et~al.(2019)Devlin, Chang, Lee, and Toutanova]{devlin2019bert}
Devlin, J., Chang, M.-W., Lee, K., and Toutanova, K.
\newblock Bert: Pre-training of deep bidirectional transformers for language understanding.
\newblock In \emph{Proceedings of the 2019 Conference of the North American Chapter of the Association for Computational Linguistics: Human Language Technologies, Volume 1 (Long and Short Papers)}, pp.\  4171--4186, 2019.

\bibitem[Dubey et~al.(2024)Dubey, Jauhri, Pandey, Kadian, Al-Dahle, Letman, Mathur, Schelten, Yang, Fan, et~al.]{dubey2024llama}
Dubey, A., Jauhri, A., Pandey, A., Kadian, A., Al-Dahle, A., Letman, A., Mathur, A., Schelten, A., Yang, A., Fan, A., et~al.
\newblock The llama 3 herd of models.
\newblock \emph{arXiv preprint arXiv:2407.21783}, 2024.

\bibitem[Gao et~al.(2023)Gao, Xiong, Gao, Jia, Pan, Bi, Dai, Sun, and Wang]{gao2023retrieval}
Gao, Y., Xiong, Y., Gao, X., Jia, K., Pan, J., Bi, Y., Dai, Y., Sun, J., and Wang, H.
\newblock Retrieval-augmented generation for large language models: A survey.
\newblock \emph{arXiv preprint arXiv:2312.10997}, 2023.

\bibitem[Gou et~al.(2021)Gou, Yu, Maybank, and Tao]{gou2021knowledge}
Gou, J., Yu, B., Maybank, S.~J., and Tao, D.
\newblock Knowledge distillation: A survey.
\newblock \emph{International Journal of Computer Vision}, 129:\penalty0 1789--1819, 2021.

\bibitem[Gupta et~al.(2018)Gupta, Chinnakotla, and Shrivastava]{gupta2018retrieve}
Gupta, V., Chinnakotla, M., and Shrivastava, M.
\newblock Retrieve and re-rank: A simple and effective ir approach to simple question answering over knowledge graphs.
\newblock In \emph{Proceedings of the First Workshop on Fact Extraction and VERification (FEVER)}, pp.\  22--27, 2018.

\bibitem[Guu et~al.(2020)Guu, Lee, Tung, Pasupat, and Chang]{guu2020retrieval}
Guu, K., Lee, K., Tung, Z., Pasupat, P., and Chang, M.
\newblock Retrieval augmented language model pre-training.
\newblock In \emph{International conference on machine learning}, pp.\  3929--3938. PMLR, 2020.

\bibitem[Izacard et~al.(2021)Izacard, Caron, Hosseini, Riedel, Bojanowski, Joulin, and Grave]{izacard2021towards}
Izacard, G., Caron, M., Hosseini, L., Riedel, S., Bojanowski, P., Joulin, A., and Grave, E.
\newblock Towards unsupervised dense information retrieval with contrastive learning.
\newblock \emph{arXiv preprint arXiv:2112.09118}, 2021.

\bibitem[Jaiswal et~al.(2020)Jaiswal, Babu, Zadeh, Banerjee, and Makedon]{jaiswal2020survey}
Jaiswal, A., Babu, A.~R., Zadeh, M.~Z., Banerjee, D., and Makedon, F.
\newblock A survey on contrastive self-supervised learning.
\newblock \emph{Technologies}, 9\penalty0 (1):\penalty0 2, 2020.

\bibitem[Jeong et~al.(2024)Jeong, Baek, Cho, Hwang, and Park]{jeong2024adaptive}
Jeong, S., Baek, J., Cho, S., Hwang, S.~J., and Park, J.~C.
\newblock Adaptive-rag: Learning to adapt retrieval-augmented large language models through question complexity.
\newblock \emph{arXiv preprint arXiv:2403.14403}, 2024.

\bibitem[Jiang et~al.(2023)Jiang, Sablayrolles, Mensch, Bamford, Chaplot, Casas, Bressand, Lengyel, Lample, Saulnier, et~al.]{jiang2023mistral}
Jiang, A.~Q., Sablayrolles, A., Mensch, A., Bamford, C., Chaplot, D.~S., Casas, D. d.~l., Bressand, F., Lengyel, G., Lample, G., Saulnier, L., et~al.
\newblock Mistral 7b.
\newblock \emph{arXiv preprint arXiv:2310.06825}, 2023.

\bibitem[Joshi et~al.(2017)Joshi, Choi, Weld, and Zettlemoyer]{joshi2017triviaqa}
Joshi, M., Choi, E., Weld, D.~S., and Zettlemoyer, L.
\newblock Triviaqa: A large scale distantly supervised challenge dataset for reading comprehension.
\newblock In \emph{Proceedings of the 55th Annual Meeting of the Association for Computational Linguistics (Volume 1: Long Papers)}, pp.\  1601--1611, 2017.

\bibitem[Karpukhin et~al.(2020)Karpukhin, Oguz, Min, Lewis, Wu, Edunov, Chen, and Yih]{karpukhin2020dense}
Karpukhin, V., Oguz, B., Min, S., Lewis, P., Wu, L., Edunov, S., Chen, D., and Yih, W.-t.
\newblock Dense passage retrieval for open-domain question answering.
\newblock In \emph{Proceedings of the 2020 Conference on Empirical Methods in Natural Language Processing (EMNLP)}, pp.\  6769--6781, 2020.

\bibitem[Ke et~al.(2024)Ke, Kong, Li, Zhang, Mei, and Bendersky]{ke-etal-2024-bridging}
Ke, Z., Kong, W., Li, C., Zhang, M., Mei, Q., and Bendersky, M.
\newblock Bridging the preference gap between retrievers and {LLM}s.
\newblock In Ku, L.-W., Martins, A., and Srikumar, V. (eds.), \emph{Proceedings of the 62nd Annual Meeting of the Association for Computational Linguistics (Volume 1: Long Papers)}, pp.\  10438--10451, Bangkok, Thailand, August 2024. Association for Computational Linguistics.
\newblock \doi{10.18653/v1/2024.acl-long.562}.
\newblock URL \url{https://aclanthology.org/2024.acl-long.562/}.

\bibitem[Koo et~al.(2024)Koo, Kim, and Hwang]{koo2024optimizing}
Koo, H., Kim, M., and Hwang, S.~J.
\newblock Optimizing query generation for enhanced document retrieval in rag.
\newblock \emph{arXiv preprint arXiv:2407.12325}, 2024.

\bibitem[Kotonya \& Toni(2020)Kotonya and Toni]{kotonya2020explainable}
Kotonya, N. and Toni, F.
\newblock Explainable automated fact-checking for public health claims.
\newblock In \emph{Proceedings of the 2020 Conference on Empirical Methods in Natural Language Processing (EMNLP)}, pp.\  7740--7754, 2020.

\bibitem[Kwiatkowski et~al.(2019)Kwiatkowski, Palomaki, Redfield, Collins, Parikh, Alberti, Epstein, Polosukhin, Devlin, Lee, et~al.]{kwiatkowski2019natural}
Kwiatkowski, T., Palomaki, J., Redfield, O., Collins, M., Parikh, A., Alberti, C., Epstein, D., Polosukhin, I., Devlin, J., Lee, K., et~al.
\newblock Natural questions: a benchmark for question answering research.
\newblock \emph{Transactions of the Association for Computational Linguistics}, 7:\penalty0 453--466, 2019.

\bibitem[Kwon et~al.(2023)Kwon, Li, Zhuang, Sheng, Zheng, Yu, Gonzalez, Zhang, and Stoica]{kwon2023efficient}
Kwon, W., Li, Z., Zhuang, S., Sheng, Y., Zheng, L., Yu, C.~H., Gonzalez, J.~E., Zhang, H., and Stoica, I.
\newblock Efficient memory management for large language model serving with pagedattention.
\newblock In \emph{Proceedings of the ACM SIGOPS 29th Symposium on Operating Systems Principles}, 2023.

\bibitem[Lewis et~al.(2020)Lewis, Perez, Piktus, Petroni, Karpukhin, Goyal, K{\"u}ttler, Lewis, Yih, Rockt{\"a}schel, et~al.]{lewis2020retrieval}
Lewis, P., Perez, E., Piktus, A., Petroni, F., Karpukhin, V., Goyal, N., K{\"u}ttler, H., Lewis, M., Yih, W.-t., Rockt{\"a}schel, T., et~al.
\newblock Retrieval-augmented generation for knowledge-intensive nlp tasks.
\newblock \emph{Advances in Neural Information Processing Systems}, 33:\penalty0 9459--9474, 2020.

\bibitem[Li et~al.(2024)Li, Verga, Sen, Yang, Viswanathan, Lewis, Watanabe, and Su]{li2024alr}
Li, H., Verga, P., Sen, P., Yang, B., Viswanathan, V., Lewis, P., Watanabe, T., and Su, Y.
\newblock Alr: A retrieve-then-reason framework for long-context question answering.
\newblock \emph{arXiv preprint arXiv:2410.03227}, 2024.

\bibitem[Liu et~al.(2023)Liu, Wang, and Zha]{liu2023fingpt}
Liu, X.-Y., Wang, G., and Zha, D.
\newblock Fingpt: Democratizing internet-scale data for financial large language models.
\newblock \emph{arXiv preprint arXiv:2307.10485}, 2023.

\bibitem[Liu et~al.(2024)Liu, Ping, Roy, Xu, Lee, Shoeybi, and Catanzaro]{liu2024chatqa}
Liu, Z., Ping, W., Roy, R., Xu, P., Lee, C., Shoeybi, M., and Catanzaro, B.
\newblock Chatqa: Surpassing gpt-4 on conversational qa and rag.
\newblock \emph{arXiv preprint arXiv:2401.10225}, 2024.

\bibitem[Loshchilov \& Hutter(2018)Loshchilov and Hutter]{loshchilov2018decoupled}
Loshchilov, I. and Hutter, F.
\newblock Decoupled weight decay regularization.
\newblock In \emph{International Conference on Learning Representations}, 2018.

\bibitem[Mallen et~al.(2023)Mallen, Asai, Zhong, Das, Khashabi, and Hajishirzi]{mallen2023not}
Mallen, A.~T., Asai, A., Zhong, V., Das, R., Khashabi, D., and Hajishirzi, H.
\newblock When not to trust language models: Investigating effectiveness of parametric and non-parametric memories.
\newblock In \emph{The 61st Annual Meeting Of The Association For Computational Linguistics}, 2023.

\bibitem[Manzoor \& Jannach(2022)Manzoor and Jannach]{manzoor2022towards}
Manzoor, A. and Jannach, D.
\newblock Towards retrieval-based conversational recommendation.
\newblock \emph{Information Systems}, 109:\penalty0 102083, 2022.

\bibitem[Min et~al.(2024)Min, Gururangan, Wallace, Shi, Hajishirzi, Smith, and Zettlemoyer]{min2024silo}
Min, S., Gururangan, S., Wallace, E., Shi, W., Hajishirzi, H., Smith, N.~A., and Zettlemoyer, L.
\newblock {SILO} language models: Isolating legal risk in a nonparametric datastore.
\newblock In \emph{The Twelfth International Conference on Learning Representations}, 2024.
\newblock URL \url{https://openreview.net/forum?id=ruk0nyQPec}.

\bibitem[Minaee et~al.(2024)Minaee, Mikolov, Nikzad, Chenaghlu, Socher, Amatriain, and Gao]{minaee2024large}
Minaee, S., Mikolov, T., Nikzad, N., Chenaghlu, M., Socher, R., Amatriain, X., and Gao, J.
\newblock Large language models: A survey.
\newblock \emph{arXiv preprint arXiv:2402.06196}, 2024.

\bibitem[Nian et~al.(2024)Nian, Peng, Wang, and Fang]{nian2024w}
Nian, J., Peng, Z., Wang, Q., and Fang, Y.
\newblock W-rag: Weakly supervised dense retrieval in rag for open-domain question answering.
\newblock \emph{arXiv preprint arXiv:2408.08444}, 2024.

\bibitem[Robertson et~al.(2009)Robertson, Zaragoza, et~al.]{robertson2009probabilistic}
Robertson, S., Zaragoza, H., et~al.
\newblock The probabilistic relevance framework: Bm25 and beyond.
\newblock \emph{Foundations and Trends{\textregistered} in Information Retrieval}, 3\penalty0 (4):\penalty0 333--389, 2009.

\bibitem[Serouis \& S{\`e}des(2024)Serouis and S{\`e}des]{serouis2024exploring}
Serouis, I.~M. and S{\`e}des, F.
\newblock Exploring large language models for bias mitigation and fairness.
\newblock In \emph{1st International Workshop on AI Governance (AIGOV) in conjunction with the Thirty-Third International Joint Conference on Artificial Intelligence}, 2024.

\bibitem[Shao et~al.(2024)Shao, He, Asai, Shi, Dettmers, Min, Zettlemoyer, and Koh]{shao2024scaling}
Shao, R., He, J., Asai, A., Shi, W., Dettmers, T., Min, S., Zettlemoyer, L., and Koh, P.~W.
\newblock Scaling retrieval-based language models with a trillion-token datastore.
\newblock \emph{arXiv preprint arXiv:2407.12854}, 2024.

\bibitem[Shi et~al.(2023)Shi, Min, Yasunaga, Seo, James, Lewis, Zettlemoyer, and Yih]{shi2023replug}
Shi, W., Min, S., Yasunaga, M., Seo, M., James, R., Lewis, M., Zettlemoyer, L., and Yih, W.-t.
\newblock Replug: Retrieval-augmented black-box language models.
\newblock \emph{arXiv preprint arXiv:2301.12652}, 2023.

\bibitem[Wang et~al.(2024{\natexlab{a}})Wang, Wan, Sun, Chen, and Ar{\i}k]{wang2024astute}
Wang, F., Wan, X., Sun, R., Chen, J., and Ar{\i}k, S.~{\"O}.
\newblock Astute rag: Overcoming imperfect retrieval augmentation and knowledge conflicts for large language models.
\newblock \emph{arXiv preprint arXiv:2410.07176}, 2024{\natexlab{a}}.

\bibitem[Wang et~al.(2022)Wang, Yang, Huang, Jiao, Yang, Jiang, Majumder, and Wei]{wang2022text}
Wang, L., Yang, N., Huang, X., Jiao, B., Yang, L., Jiang, D., Majumder, R., and Wei, F.
\newblock Text embeddings by weakly-supervised contrastive pre-training.
\newblock \emph{arXiv preprint arXiv:2212.03533}, 2022.

\bibitem[Wang et~al.(2024{\natexlab{b}})Wang, Ma, Feng, Zhang, Yang, Zhang, Chen, Tang, Chen, Lin, et~al.]{wang2024survey}
Wang, L., Ma, C., Feng, X., Zhang, Z., Yang, H., Zhang, J., Chen, Z., Tang, J., Chen, X., Lin, Y., et~al.
\newblock A survey on large language model based autonomous agents.
\newblock \emph{Frontiers of Computer Science}, 18\penalty0 (6):\penalty0 1--26, 2024{\natexlab{b}}.

\bibitem[Wang et~al.(2023)Wang, Fei, Leng, and Li]{wang2023does}
Wang, X., Fei, Y., Leng, Z., and Li, C.
\newblock Does role-playing chatbots capture the character personalities? assessing personality traits for role-playing chatbots.
\newblock \emph{arXiv preprint arXiv:2310.17976}, 2023.

\bibitem[Wang et~al.(2024{\natexlab{c}})Wang, Wang, Gao, Zhang, Wu, Xu, Shi, Wang, Li, Qian, et~al.]{wang2024searching}
Wang, X., Wang, Z., Gao, X., Zhang, F., Wu, Y., Xu, Z., Shi, T., Wang, Z., Li, S., Qian, Q., et~al.
\newblock Searching for best practices in retrieval-augmented generation.
\newblock In \emph{Proceedings of the 2024 Conference on Empirical Methods in Natural Language Processing}, pp.\  17716--17736, 2024{\natexlab{c}}.

\bibitem[Wang et~al.(2024{\natexlab{d}})Wang, Wang, Le, Zheng, Mishra, Perot, Zhang, Mattapalli, Taly, Shang, et~al.]{wang2024speculative}
Wang, Z., Wang, Z., Le, L., Zheng, H.~S., Mishra, S., Perot, V., Zhang, Y., Mattapalli, A., Taly, A., Shang, J., et~al.
\newblock Speculative rag: Enhancing retrieval augmented generation through drafting.
\newblock \emph{arXiv preprint arXiv:2407.08223}, 2024{\natexlab{d}}.

\bibitem[Wang et~al.(2024{\natexlab{e}})Wang, Asai, Yu, Xu, Xie, Neubig, and Fried]{wang2024coderag}
Wang, Z.~Z., Asai, A., Yu, X.~V., Xu, F.~F., Xie, Y., Neubig, G., and Fried, D.
\newblock Coderag-bench: Can retrieval augment code generation?
\newblock \emph{arXiv preprint arXiv:2406.14497}, 2024{\natexlab{e}}.

\bibitem[Wenzek et~al.(2019)Wenzek, Lachaux, Conneau, Chaudhary, Guzm{\'a}n, Joulin, and Grave]{wenzek2019ccnet}
Wenzek, G., Lachaux, M.-A., Conneau, A., Chaudhary, V., Guzm{\'a}n, F., Joulin, A., and Grave, E.
\newblock Ccnet: Extracting high quality monolingual datasets from web crawl data.
\newblock \emph{arXiv preprint arXiv:1911.00359}, 2019.

\bibitem[Wu \& Cao(2024)Wu and Cao]{wu2024llm}
Wu, M. and Cao, S.
\newblock Llm-augmented retrieval: Enhancing retrieval models through language models and doc-level embedding.
\newblock \emph{arXiv preprint arXiv:2404.05825}, 2024.

\bibitem[Wu et~al.(2024)Wu, Xie, Chen, Zhu, Zhang, and Xiao]{wu2024easily}
Wu, S., Xie, J., Chen, J., Zhu, T., Zhang, K., and Xiao, Y.
\newblock How easily do irrelevant inputs skew the responses of large language models?
\newblock \emph{arXiv preprint arXiv:2404.03302}, 2024.

\bibitem[Xiong et~al.(2020)Xiong, Xiong, Li, Tang, Liu, Bennett, Ahmed, and Overwijk]{xiong2020approximate}
Xiong, L., Xiong, C., Li, Y., Tang, K.-F., Liu, J., Bennett, P., Ahmed, J., and Overwijk, A.
\newblock Approximate nearest neighbor negative contrastive learning for dense text retrieval.
\newblock \emph{arXiv preprint arXiv:2007.00808}, 2020.

\bibitem[Xu et~al.(2023)Xu, Ping, Wu, McAfee, Zhu, Liu, Subramanian, Bakhturina, Shoeybi, and Catanzaro]{xu2023retrieval}
Xu, P., Ping, W., Wu, X., McAfee, L., Zhu, C., Liu, Z., Subramanian, S., Bakhturina, E., Shoeybi, M., and Catanzaro, B.
\newblock Retrieval meets long context large language models.
\newblock \emph{arXiv preprint arXiv:2310.03025}, 2023.

\bibitem[Yu et~al.(2024)Yu, Ping, Liu, Wang, You, Zhang, Shoeybi, and Catanzaro]{yu2024rankrag}
Yu, Y., Ping, W., Liu, Z., Wang, B., You, J., Zhang, C., Shoeybi, M., and Catanzaro, B.
\newblock Rankrag: Unifying context ranking with retrieval-augmented generation in llms.
\newblock \emph{arXiv preprint arXiv:2407.02485}, 2024.

\bibitem[Yue et~al.(2024)Yue, Wang, Chen, Huang, and Wei]{yue2024synergistic}
Yue, S., Wang, S., Chen, W., Huang, X., and Wei, Z.
\newblock Synergistic multi-agent framework with trajectory learning for knowledge-intensive tasks.
\newblock \emph{arXiv preprint arXiv:2407.09893}, 2024.

\bibitem[Zhang et~al.(2024)Zhang, Song, Wang, Tang, Li, Zeng, Wu, Ye, Xu, Zhang, et~al.]{zhang2024raglab}
Zhang, X., Song, Y.-Z., Wang, Y., Tang, S., Li, X., Zeng, Z., Wu, Z., Ye, W., Xu, W., Zhang, Y., et~al.
\newblock Raglab: A modular and research-oriented unified framework for retrieval-augmented generation.
\newblock In \emph{Proceedings of the 2024 Conference on Empirical Methods in Natural Language Processing: System Demonstrations}, pp.\  408--418, 2024.

\bibitem[Zhao et~al.(2023)Zhao, Zhou, Li, Tang, Wang, Hou, Min, Zhang, Zhang, Dong, et~al.]{zhao2023survey}
Zhao, W.~X., Zhou, K., Li, J., Tang, T., Wang, X., Hou, Y., Min, Y., Zhang, B., Zhang, J., Dong, Z., et~al.
\newblock A survey of large language models.
\newblock \emph{arXiv preprint arXiv:2303.18223}, 2023.

\bibitem[Zhou et~al.(2022)Zhou, Li, Shang, Luo, Zhan, Hu, Zhang, Jiang, Cao, Yu, et~al.]{zhou2022hyperlink}
Zhou, J., Li, X., Shang, L., Luo, L., Zhan, K., Hu, E., Zhang, X., Jiang, H., Cao, Z., Yu, F., et~al.
\newblock Hyperlink-induced pre-training for passage retrieval in open-domain question answering.
\newblock In \emph{Proceedings of the 60th Annual Meeting of the Association for Computational Linguistics (Volume 1: Long Papers)}, pp.\  7135--7146, 2022.

\bibitem[Zhou et~al.(2024{\natexlab{a}})Zhou, Dong, Wei, and Chen]{zhou2024semi}
Zhou, J., Dong, L., Wei, F., and Chen, L.
\newblock Semi-parametric retrieval via binary token index.
\newblock \emph{arXiv preprint arXiv:2405.01924}, 2024{\natexlab{a}}.

\bibitem[Zhou et~al.(2024{\natexlab{b}})Zhou, Li, Shang, Jiang, Liu, and Chen]{zhou2024retrievalbased}
Zhou, J., Li, X., Shang, L., Jiang, X., Liu, Q., and Chen, L.
\newblock Retrieval-based disentangled representation learning with natural language supervision.
\newblock In \emph{The Twelfth International Conference on Learning Representations}, 2024{\natexlab{b}}.
\newblock URL \url{https://openreview.net/forum?id=ZlQRiFmq7Y}.

\end{thebibliography}
\bibliographystyle{icml}

%%%%%%%%%%%%%%%%%%%%%%%%%%%%%%%%%%%%%%%%%%%%%%%%%%%%%%%%%%%%%%%%%%%%%%%%%%%%%%%
%%%%%%%%%%%%%%%%%%%%%%%%%%%%%%%%%%%%%%%%%%%%%%%%%%%%%%%%%%%%%%%%%%%%%%%%%%%%%%%
% APPENDIX
%%%%%%%%%%%%%%%%%%%%%%%%%%%%%%%%%%%%%%%%%%%%%%%%%%%%%%%%%%%%%%%%%%%%%%%%%%%%%%%
%%%%%%%%%%%%%%%%%%%%%%%%%%%%%%%%%%%%%%%%%%%%%%%%%%%%%%%%%%%%%%%%%%%%%%%%%%%%%%%
\newpage
\appendix
\onecolumn
\section{Details of Datasets}
\label{appendix:dataset}

We present details of datasets as follows.

\begin{itemize}
    \item Natural Questions (NQ;~\citeauthor{kwiatkowski2019natural}, \citeyear{kwiatkowski2019natural}) is a widely used open-domain QA dataset constructed from Wikipedia. The questions originate from Google search queries, and the answers are text spans within Wikipedia passages. This dataset consists of queries with one or more answer strings, requiring RAG systems to generate responses based on factual knowledge. 
    \item TriviaQA (TQA;~\citeauthor{joshi2017triviaqa}, \citeyear{joshi2017triviaqa}) is a challenging QA dataset that comprises question-answer pairs curated by trivia enthusiasts along with independently gathered evidence documents.
    \item PubHealth~\cite{kotonya2020explainable} is a fact-checking task that focuses on verifying health claims across a variety of biomedical topics.
    \item ARC-Challenge~\cite{clark2018think} is a multiple-choice reasoning dataset consisting of science exam questions for grades 3 to 9.
\end{itemize}
\section{Details of Baseline Models}
\label{appendix:models}

The information for baseline models are listed as follows.

\subsection{Retrieval Model (IR)}
\begin{itemize}
    \item \efive~\cite{wang2022text} is a state-of-the-art dense retriever that pre-trained on millions of weakly related text pairs from the Web. The unsupervised version of this model is denoted as $\text{E5-unsup}$. This model undergoes further fine-tuning on natural language inference (NLI) datasets, as well as the Natural Questions and MS MARCO datasets, to enhance its capabilities in downstream applications. The fine-tuned version is denoted as \efive.
    
    \item \contriever~\cite{izacard2021towards} is a widely-used dense retriever pre-trained unsupervised on Wikipedia data and CCNet~\cite{wenzek2019ccnet}. The unsupervised version of this model is denoted as \contriever. It is further fine-tuned on the MS MARCO dataset to enhance its retrieval performance, with the fine-tuned version denoted as \contrieverms.
    
    \item DPR~\cite{karpukhin2020dense} is a widely used dense passage retriever initialized with a BERT-based uncased encoder~\cite{devlin2019bert}, and fine-tuned on downstream dataset. Specifically, $\text{DPR}_\text{MS}$ is fine-tuned on the MS MARCO dataset, $\text{DPR}_\text{NQ}$ on the NQ dataset, and $\text{DPR}_\text{TQA}$ on the TriviaQA dataset.
    
    \item \sidr~\cite{zhou2024semi} is a semi-parametric sparse retriever that supports using both embeddings and tokenization as index. This nature allows for in-training retrieval, where the model's parameters dynamically update while the retrieval index remains fixed. The model is initialized with a BERT-based uncased encoder~\cite{devlin2019bert} and fine-tuned exclusively on single dataset depending on the variant: \sidrms is fine-tuned on the MS MARCO dataset, \sidrnq on the NQ dataset, and \sidrtqa on the TriviaQA dataset.
\end{itemize}
All the above retrieval methods are initialized with a BERT-based encoder, which contains approximately 200 million (0.2B) parameters.

\subsection{Large Language Model (LLM)}
\begin{itemize}
    \item Llama3$_\text{8B}$~\cite{dubey2024llama} is a variant of the latest Llama3 model series with 8 billion parameters. 
    \item Llama3-Instruct$_\text{8B}$~\cite{dubey2024llama} builds upon the Llama3$_\text{8B}$ by undergoing a post-training stage in which the model is specifically tuned to follow instructions and align with human preferences to improve specific capabilities.
    \item Phi-3-mini-4k-instruct$_\text{3.8B}$~\cite{abdin2024phi} is a lightweight widely-used LLM with 3.8 billion parameters, trained on the Phi-3 dataset featuring synthetic and high-quality filtered web data, focused on reasoning and quality. 
    \item Mistral-Instruct$_\text{7B}$~\cite{jiang2023mistral}. We use Mistral-7B-Instruct-v0.3 LLM which is an instruct fine-tuned version of the Mistral-7B-v0.3.
\end{itemize}

\subsection{Retrieval-augmented Generation Framework (RAG)}
\begin{itemize}
    \item \replug~\cite{shi2023replug} is a RAG framework using GPT-3 and \contriever. The retriever is specifically trained to use the first 128 tokens of a sequence as queries, with the goal of retrieving documents that maximize the probability of generating the subsequent 128 tokens when these retrieved documents are prepended to the query.
    
    % Our reproduction of \replug utilizes the same LLM, Llama3$_\text{8B}$, the retriever \sidrms, datastore, and prompt as \openrag. The primary distinction between our reproduction of \replug and \openrag lies in the training objective where \openrag employs contrastive learning and our \replug utilizes KL-divergence for training.

    \item \selfrag~\cite{asai2023self} is a RAG framework designed to improve response quality by enabling on-demand retrieval and incorporating self-reflection mechanisms.

    The reproductions by \citet{wang2024speculative} and \citet{zhang2024raglab}, $\textsc{Self-RAG}_\text{Mistral-7B}$ and $\textsc{Self-RAG}_\text{Llama3-8B}$ respectively, involve tuning Mistral-7B and Llama3-8B as base language models using the open-source data provided by \selfrag.
    
    Our reproduction, $\textsc{Self-RAG}_\text{Llama3-8B} + $\sidrms, utilizes the $\textsc{Self-RAG}_\text{Llama3-8B}$ checkpoint from \citet{zhang2024raglab} as LLM, while employing the same retriever \sidrms and adapting it to our downstream setup.
\end{itemize}

\section{Effectiveness of RAG Scores on Task Accuracy}
\label{appendix:ragscores}

\begin{table*}[ht]
  \centering
  \caption{Results of RAG framework using top-1 and top-10 documents in context, sorted by retrieval relevance and RAG scores.}
% Table generated by Excel2LaTeX from sheet 'appendix'
\scalebox{0.8}{
\begin{tabular}{lcccccccc}
\toprule
\textbf{Task Type ($\rightarrow$)} & \multicolumn{4}{c}{\textbf{Free-form}} & \multicolumn{4}{c}{\textbf{Closed-set}} \\
\cmidrule(lr){2-5}\cmidrule(lr){6-9}
\textbf{\quad Dataset ($\rightarrow$)} & \multicolumn{2}{c}{\textbf{NQ}} & \multicolumn{2}{c}{\textbf{TriviaQA}} & \multicolumn{2}{c}{\textbf{PubHealth}} & \multicolumn{2}{c}{\textbf{ARC-C}} \\
\cmidrule(lr){2-3}\cmidrule(lr){4-5}\cmidrule(lr){6-7}\cmidrule(lr){8-9}
\cmidrule{2-3}\textbf{Method ($\downarrow$) \quad Metrics ($\rightarrow$)} & \textbf{1-doc} & \textbf{10-doc} & \textbf{1-doc} & \textbf{10-doc} & \textbf{1-doc} & \textbf{10-doc} & \textbf{1-doc} & \textbf{10-doc} \\
\midrule
\quad Llama3$_\text{8B}$ + \sidrms (doc with top relevance) & 49.1 & 51.4 & 65.3 & 67.2 & 65.2 & 67.4 & 58.1 & 57.3 \\
\quad Llama3$_\text{8B}$ + \sidrms (doc with top RAG scores) & 85.1 & 76.2 & 88.7 & 84.2 & 87.4 & 77.4 & 95.6 & 83.6 \\
\bottomrule
\end{tabular}}
\label{tab:ragscores}%
\end{table*}

Given that our learning is based on using the RAG score as an indicator to identify positive and negative documents, we now investigate whether using documents with higher RAG scores leads to improved RAG response quality. For each dataset, we sample 1k samples from training split. For each query, we retrieve the top 100 documents, and then perform the RAG pipeline using only the top-1 and top-10 documents, sorted by retrieval relevance and RAG scores, respectively. The results, shown in Table~\ref{tab:ragscores}, indicate that RAG scores are indicative of the final accuracy of the RAG framework. Furthermore, the high accuracy achieved using top RAG scores documents suggests that the datastore holds significant untapped potential, which current retrieval strategies have not yet fully exploited. 

To our knowledge, using RAG scores to identify positives and negatives is a rough yet resource-efficient solution that could cover most existing knowledge-intensive tasks, aligning with their evaluation metrics that often utilize string matching. However, it may not be suitable for long-form generation, which requires different evaluation strategies. We believe it is possible to customize the identification of positive and negative examples based on the specific needs of each task. Ideally, if computational cost is not a concern or resources are sufficient, a strong proprietary LLM like GPT-4 can be used for contrastive identification on-the-fly.

Here are some additional observations: RAG scores are generally more indicative when using single document in context, likely because they are computed in this manner, ensuring more consistent evaluations. Furthermore, the improved performance seen in Table~\ref{tab:ragscores} compared to our main experiments may be attributed to the LLM having been pretrained on the training split of these datasets.

\section{Revisiting Semi-parametric Disentangled Retriever (\sidr)}
\label{appendix:sidr}

\newcommand{\innerproduct}[2]{\langle #1, #2 \rangle}

Our work adopts the recently proposed retriever \sidr as the backbone for two main reasons. First, it supports the use of a non-parametric index, which enables in-training retrieval when the retriever's parameters change dynamically. Second, evaluating retriever checkpoints can be resource-intensive, as it requires embedding a large datastore with each new checkpoint. \sidr offers late parametric techniques that reduce this evaluation process from a full day on our resource to just a few minutes, significantly accelerating our research.

\begin{figure*}[ht]
\begin{center}
\includegraphics[width=0.98\textwidth]{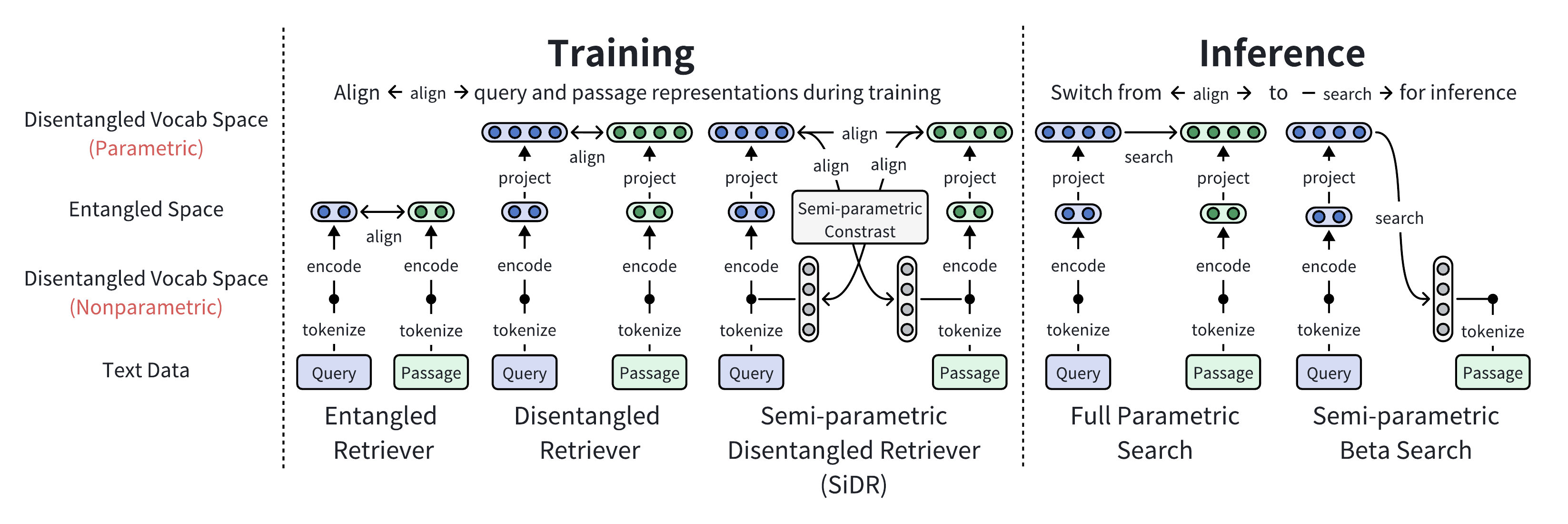}
\vspace{-0.5cm}
\caption{Illustration of semi-parametric disentangled retriever (\sidr) framework, adapted from~\citet{zhou2024semi}.}
\label{fig:sidr}
\end{center}
\vspace{-0.2cm}
\end{figure*}

\sidr~\cite{zhou2024retrievalbased,zhou2024semi} is a sparse disentangled retriever (also known as a sparse lexical retriever) that encodes text chunks into a $|V|$-dimensional sparse representation, where each dimension represents the importance of a token within the language model vocabulary $V$. \sidr is then trained to align the $|V|$-dimensional parametric embedding, denoted as $V_\theta(x)$,  with the $|V|$-dimensional bag-of-tokens representation, denoted as $V_\text{BoT}(x)$.

At downstream, a parametric query embedding $V_\theta(q)$ can perform search on both an embedding-based index $V_{\theta}(\mathcal{D})$ and a bag-of-tokens index $V_\text{BoT}(\mathcal{D})$, which leads to three distinct search schemes:
\begin{itemize}
    \item \textbf{Full parametric search} utilizes a \textbf{parametric index} $V_{\theta}(\mathcal{D})$, which relies on embeddings derived from a neural encoder for the datastore. The relevance is defined as the inner product of the embeded query and embeded datastore: 
    \begin{equation} f_{\theta}(q, \mathcal{D}) = \innerproduct{V_{\theta}(q)}{V_{\theta}(\mathcal{D})} \nonumber \end{equation} 
    This is the common indexing process for neural retrieval systems, which are effective but involve higher costs and longer latency for embedding the entire $\mathcal{D}$ to obtain the index $V_{\theta}(\mathcal{D})$.

    \item \textbf{Semi-parametric beta search} leverages a \textbf{non-parametric index} $V_{\text{BoT}}(\mathcal{D})$ based on BoT representations of the datastore, which are constructed solely by a tokenizer. The relevance is defined as: 
    \begin{equation}
    f_{\beta}(q, \mathcal{D}) = \innerproduct{V_{\theta}(q)}{V_{\text{BoT}}(\mathcal{D})} \nonumber 
    \end{equation} 

    \item \textbf{Late parametric with top-m re-rank} is a search pipeline that starts search with a non-parametric index to retrieve top-$m$ passages, denote as $\mathcal{D}_{m}$, and then on-the-fly embeds them for re-ranking: 
    \begin{align}
    f_{\beta}(q, \mathcal{D}) = \innerproduct{V_{\theta}(q)}{V_{\text{BoT}}(\mathcal{D})} \nonumber; \quad
    f_{\theta}(q, \mathcal{D}_{m}) = \innerproduct{V_{\theta}(q)}{V_{\theta}(\mathcal{D}_{m})} \nonumber
    \end{align}

\end{itemize}

In our framework, we primarily utilize the late parametric techniques provided by \sidr. For in-training retrieval, we use late parametric with top-20 re-ranking. For checkpoint evaluation and inspection in the ablation study, we use late parametric with top-100 re-ranking to accelerate results while managing limited resources. In our main experiments, we use full parametric search.
\section{Late Parametric vs. Periodic Re-indexing}
\label{appendix:latepara}

\newcommand{\reindex}{\openrag (w/ re-index)\xspace}
\newcommand{\rerank}{\openrag (w/o re-rank)\xspace}

A key distinction between our work and prior practices lies in our use of the late parametric mechanism to avoid re-indexing during training. 
In this section, we systematically evaluate these in-training retrieval approaches.

\textbf{Baseline.} We present ablation studies on different in-training retrieval approaches: 
(i)~\underline{\openrag} employs the late parametric method as proposed in \sidr, which uses a bag-of-token index for first-stage retrieval and re-ranks the top-20 documents on-the-fly using up-to-date parameters.
(ii) \underline{\rerank} employs the bag-of-token index for retrieval, similar to the late parametric method but without the re-ranking process. This setup aims to assess the costs associated with re-ranking during training.
(iii) \underline{\reindex} involves periodic re-indexing using the most recently built but outdated index for retrieval, an in-training retrieval method that commonly used in prior studies. In this setup, we employ $\text{DPR}_{\text{MS}}$ as the initial retriever. We avoid using \sidrms, which has high-dimensional embeddings of 30,522, in stark contrast to DPR's 768 dimensions. This significant discrepancy prevents our GPU cards from allocating the parametric index for \sidrms, although they manage DPR effectively.

\textbf{Training.} All models undergo the similar training pipeline: they are trained for 80 epochs with the first 40 epochs as a warm-up and the last 40 conducting in-training retrieval. They differ only in their in-training retrieval strategies: both \openrag and \rerank do not require re-indexing; \reindex requires rebuilding index at every 15 epochs (around 5k steps), a rebuild interval commonly used in previous research~\cite{xiong2020approximate}, resulting in a total of three rebuilds.

\begin{figure*}[ht]
\begin{center}
\includegraphics[width=0.8\textwidth]{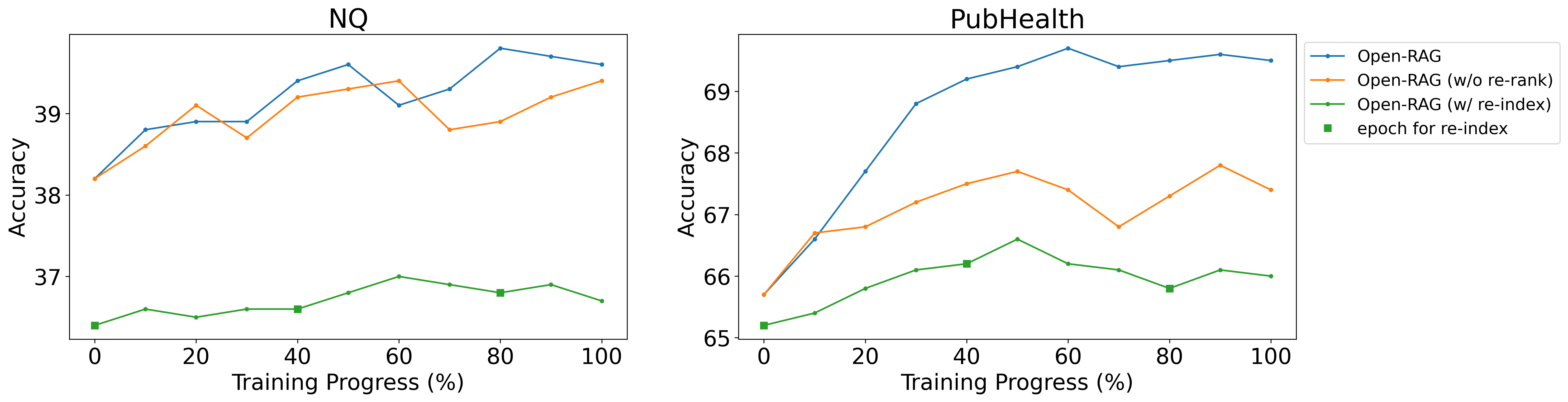}
\caption{RAG accuracy of different in-training retrieval approaches.}
\label{fig:latepara}
\end{center}
\end{figure*}

\textbf{Results.}  We present the RAG accuracy on NQ and PubHealth test splits during in-training retrieval, with results reported every four epochs, as depicted in Figure~\ref{fig:latepara}. For the re-ranking setup, significant improvements are observed in the PubHealth data when re-ranking is employed, whereas the NQ dataset shows only minor improvements. Given that the costs associated with re-ranking are manageable in our setup, we continue to implement it. Regarding re-indexing, our findings indicate that despite requiring significant time and resources, it fails to yield improvements comparable to those of the late parametric approach and significantly lags behind. We attribute this to index staleness, where query embeddings must optimize against outdated document embeddings, rendering the learning process less effective. On the other hand, as presented in the study by \citet{zhou2024semi}, by re-ranking the top-20 retrieved documents, the late parametric method can recover more than 90\% of the performance of a full parametric search across different tasks, representing a minor compromise. This also partially explains why the late parametric approach outperforms periodic re-indexing.

\section{Inconsistencies between IR and RAG Scenarios}
\label{appendix:gap}

\subsection{Performance Changes in IR Scenarios after Tuning}

\begin{table}[ht]\small
\centering
\small
\caption{Performance changes before and after tuning the retriever using the \openrag approach.}
\begin{tabular}{lcccc}
\cmidrule{1-5}
\textbf{Dataset ($\rightarrow$)} & \multicolumn{2}{c}{\textbf{NQ}} & \multicolumn{2}{c}{\textbf{TriviaQA}} \\
\cmidrule(lr){2-3}\cmidrule(lr){4-5}
\textbf{Method ($\downarrow$) \quad Metrics ($\rightarrow$)}\unboldmath{} & \textbf{IR} & \textbf{RAG} & \textbf{IR} & \textbf{RAG} \\
\midrule
\quad Llama3$_\text{8B}$ + \sidrms & 39.1 & 34.4 & 56.1 & 62.0 \\
\quad Llama3$_\text{8B}$ + \openrag (\sidrms) & 40.8 (\increase{1.7}) & 39.8 (\increase{5.4}) & 53.9  (\decrease{2.2}) & 65.8  (\increase{3.8}) \\
\midrule
\quad Llama3$_\text{8B}$ + \sidrnq & 49.5 & 42.7 & -- & -- \\
\quad Llama3$_\text{8B}$ + \openrag (\sidrnq) & 47.1  (\decrease{2.4}) & 44.1  (\increase{1.4}) & -- & -- \\
\bottomrule
\end{tabular}
\label{tab:gap}
\end{table}

We evaluate the performance of our retriever in both IR and RAG scenarios before and after tuning. In IR scenarios, we measure top-1 retrieval accuracy by checking whether the top-1 retrieved document contains the answer. In RAG scenarios, we measure accuracy using a single document in the context window, evaluating whether the generated response contains the correct answer.

Our results indicate that while \openrag tunes the retriever to improve RAG performance, it results in inconsistent performance on traditional IR performance, with some degradation observed on certain datasets. This highlights a long-standing issue in the IR evaluation pipeline: a document containing the answer does not necessarily imply that it effectively addresses the query, and conversely, a document not containing the answer does not mean it is irrelevant or unhelpful. 

Our conclusion also aligns with the findings and observations of other research. \citet{cuconasu2024power} find that including more answer-containing documents in the context negatively impacts RAG performance. Similarly, \citet{nian2024w} observe that traditional relevance definitions for IR tasks do not enhance RAG response quality. Additional research emphasizes the need for further learning to bridge the preference gap~\cite{ke-etal-2024-bridging} or re-ranking~\cite{yu2024rankrag} for off-the-shelf retrievers to improve RAG performance.

\subsection{Case Study}
In this section, we present a case study using the NQ dataset where each query has a list of answer strings. This case study is designed to further explore the inconsistency issues inherent in RAG implementations. We specifically examine two scenarios: (i) cases where the retrieved document contains the correct answer but fails to produce the correct RAG output, and (ii) instances where the retrieved document does not directly address the query, yet the RAG model manages to generate the correct answer nonetheless. To enhance our analysis, we also ask GPT-4 to judge whether the documents address the question, helping readers quickly grasp the key issue.

\begin{figure*}[ht]
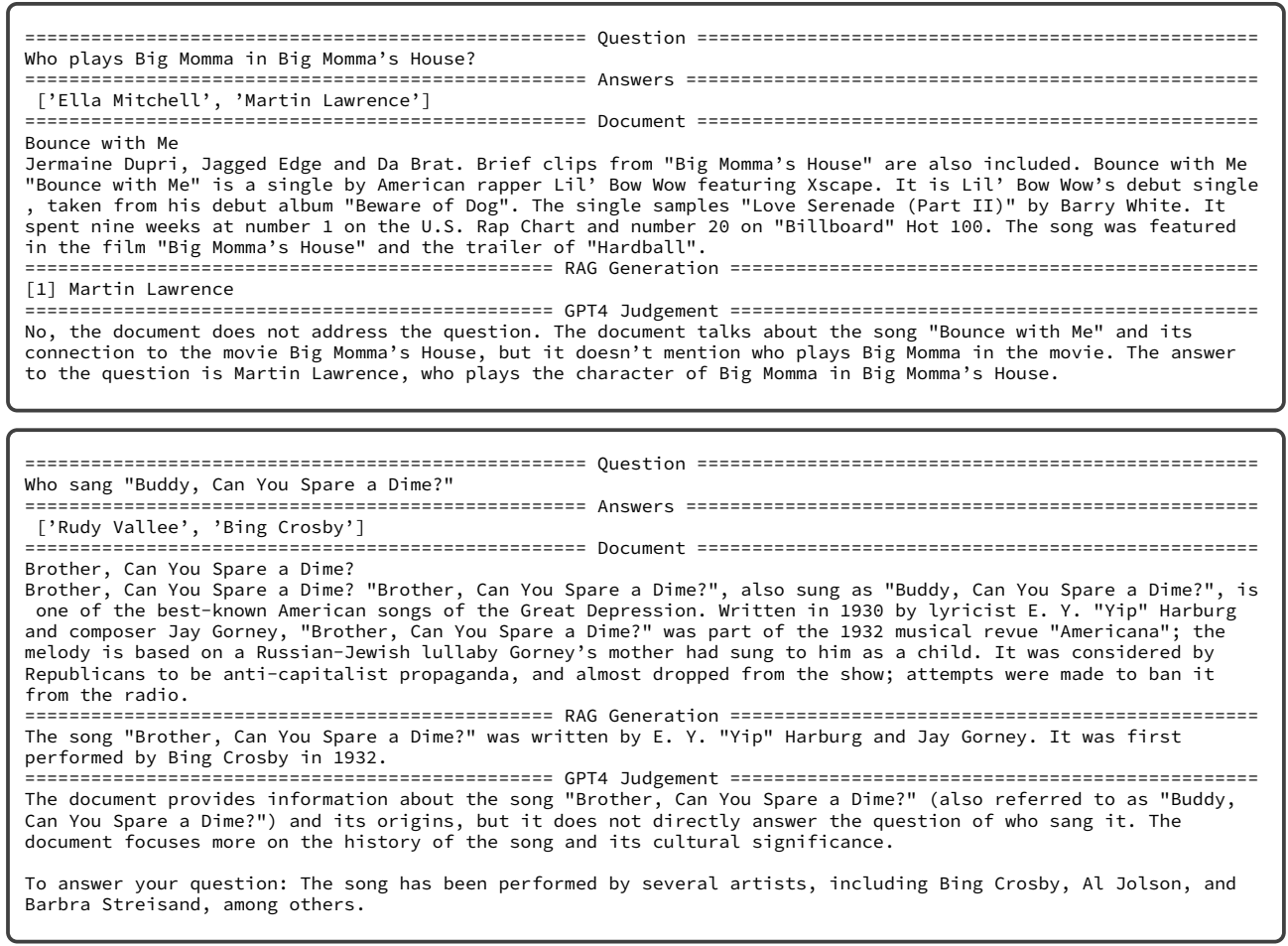

\begin{tcolorbox}[left=1mm,right=1mm,top=0.mm, bottom=0mm,colback=white]
\begin{lstlisting}[style=demo, breaklines=true, breakindent=0pt]
=================================================== Question =================================================== 
Who plays Big Momma in Big Momma's House?
=================================================== Answers ==================================================== 
 ['Ella Mitchell', 'Martin Lawrence']
=================================================== Document =================================================== 
Bounce with Me
Jermaine Dupri, Jagged Edge and Da Brat. Brief clips from "Big Momma's House" are also included. Bounce with Me "Bounce with Me" is a single by American rapper Lil' Bow Wow featuring Xscape. It is Lil' Bow Wow's debut single, taken from his debut album "Beware of Dog". The single samples "Love Serenade (Part II)" by Barry White. It spent nine weeks at number 1 on the U.S. Rap Chart and number 20 on "Billboard" Hot 100. The song was featured in the film "Big Momma's House" and the trailer of "Hardball". 
================================================ RAG Generation ================================================ 
[1] Martin Lawrence
================================================ GPT4 Judgement ================================================ 
No, the document does not address the question. The document talks about the song "Bounce with Me" and its connection to the movie Big Momma's House, but it doesn't mention who plays Big Momma in the movie. The answer to the question is Martin Lawrence, who plays the character of Big Momma in Big Momma's House.
\end{lstlisting}
\end{tcolorbox}

\begin{tcolorbox}[left=1mm,right=1mm,top=0.mm, bottom=0mm,colback=white]
\begin{lstlisting}[style=demo, breaklines=true, breakindent=0pt]
=================================================== Question =================================================== 
Who sang "Buddy, Can You Spare a Dime?"
=================================================== Answers ==================================================== 
 ['Rudy Vallee', 'Bing Crosby']
=================================================== Document =================================================== 
Brother, Can You Spare a Dime? 
Brother, Can You Spare a Dime? "Brother, Can You Spare a Dime?", also sung as "Buddy, Can You Spare a Dime?", is one of the best-known American songs of the Great Depression. Written in 1930 by lyricist E. Y. "Yip" Harburg and composer Jay Gorney, "Brother, Can You Spare a Dime?" was part of the 1932 musical revue "Americana"; the melody is based on a Russian-Jewish lullaby Gorney's mother had sung to him as a child. It was considered by Republicans to be anti-capitalist propaganda, and almost dropped from the show; attempts were made to ban it from the radio.
================================================ RAG Generation ================================================ 
The song "Brother, Can You Spare a Dime?" was written by E. Y. "Yip" Harburg and Jay Gorney. It was first performed by Bing Crosby in 1932.
================================================ GPT4 Judgement ================================================ 
The document provides information about the song "Brother, Can You Spare a Dime?" (also referred to as "Buddy, Can You Spare a Dime?") and its origins, but it does not directly answer the question of who sang it. The document focuses more on the history of the song and its cultural significance.

To answer your question: The song has been performed by several artists, including Bing Crosby, Al Jolson, and Barbra Streisand, among others.
\end{lstlisting}
\end{tcolorbox}

\caption{Example of RAG output correct answers while the retrieved document does not contain the correct answer, nor does GPT-4 consider the document address the question.}
\label{fig:case1}
\end{figure*}

In Figure~\ref{fig:case1}, we present examples where RAG outputs the correct answer, even though the retrieved document neither contains the answer nor is considered to address the question by GPT-4. In both cases, the document fails to provide the correct answer or relevant clues, yet RAG is still able to generate the correct response. We believe this is a common phenomenon, as LLMs possess a wealth of internal knowledge, particularly for public knowledge questions. In general, an incorrect or imperfect retrieved document is insufficient to mislead the LLM into producing an incorrect output.

\begin{figure*}[ht]
\begin{tcolorbox}[left=1mm,right=1mm,top=0.mm, bottom=0mm,colback=white]
\begin{lstlisting}[style=demo, breaklines=true, breakindent=0pt]
=================================================== Question =================================================== 
When was The Hitchhiker's Guide to the Galaxy made?
=================================================== Answers ==================================================== 
['1978', '2005']
=================================================== Document =================================================== 
The Hitchhiker's Guide to the Galaxy (radio series)
The original series was the recipient of a number of awards including the Imperial Tobacco Award (1978), The Sony Award (1979), The Society of Authors/Pye Awards Best Programme for Young People (1980) and the Mark Time Awards 'Grand Master Award' (Adams) and 'Hall of Fame' (1998). It was the only radio show ever to be nominated for the Hugo science fiction awards, in 1979, in the 'Best Dramatic Presentation' category. As a result of the series, Douglas Adams was inducted into the Radio Academy's Hall of Fame. In November 2003, two years after Adams's death and 23 years after the ...
 
================================================ RAG Generation ================================================ 
[1] The Hitchhiker's Guide to the Galaxy (radio series)
================================================ GPT4 Judgement ================================================ 
No, the document does not directly address the query of when The Hitchhiker's Guide to the Galaxy was made. It provides information about the radio series, its awards, and recognition but does not specify the year of its creation or first broadcast.

The original radio series of The Hitchhiker's Guide to the Galaxy first aired in 1978.
\end{lstlisting}
\end{tcolorbox}

\begin{tcolorbox}[left=1mm,right=1mm,top=0.mm, bottom=0mm,colback=white]
\begin{lstlisting}[style=demo, breaklines=true, breakindent=0pt]
=================================================== Question =================================================== 
Who ordered the Forbidden City to be built?
=================================================== Answers ==================================================== 
['Zhu Di']
=================================================== Document =================================================== 
History of the Forbidden City
The site of the Forbidden City was situated on the Imperial city during the Mongol Yuan Dynasty. After the collapse of the Yuan Dynasty, the Hongwu Emperor of the Ming Dynasty moved the capital from Beijing in the north to Nanjing in the south, and in 1369 ordered that the Yuan palaces be razed. His son Zhu Di was created Prince of Yan with his seat in Beijing. In 1402, Zhu Di usurped the throne and became the Yongle Emperor. He made Beijing a secondary capital of the Ming empire ...

================================================ RAG Generation ================================================ 
The Yongle Emperor ordered the forbidden city to be built.
================================================ GPT4 Judgement ================================================ 
Yes, the document addresses the query. It mentions that Zhu Di, who later became the Yongle Emperor, ordered the construction of the Forbidden City. The passage describes how he moved the capital to Beijing and commissioned the building of the Forbidden City in 1406.
\end{lstlisting}
\end{tcolorbox}

\caption{Example of RAG failing to output the correct answer while the retrieved document contains the correct answer or GPT-4 considers the document as addressing the question.}
\label{fig:case2}
\end{figure*}

In Figure~\ref{fig:case2}, we present examples where RAG fails to output the correct answer, even though the retrieved document contains the correct answer or GPT-4 considers the document as addressing the question. In the first case, the document does not address the query, and the LLM tends to extract key phrases, such as the title, as the response, ignoring the query. In the second case, the document contains information that addresses the query, and the LLM generates the correct answer, but the answer’s alias name is not included in the pre-defined answer candidates, leading to a failure in the RAG scenario. These inconsistencies can be driven by many factors, including the LLM, instruction prompt, evaluation metrics, and relevance. All of these factors are intertwined, and we believe that end-to-end data-driven learning is more effective than analyzing their interplay in isolation.

\newpage

\section{Case Study of RAG Labels}
\label{appendix:raglabels}

For free-form generation tasks, we assess whether the generation contains any of the given answers. For closed-set generation tasks, we measure whether the generation contains the label. Below are examples that illustrate how different generations lead to different RAG labels given the same question and answers.

\begin{figure*}[ht]
\begin{tcolorbox}[left=1mm,right=1mm,top=0.mm, bottom=0mm,colback=white]
\begin{lstlisting}[style=demo, breaklines=true, breakindent=0pt]
================================================== Question ==================================================== 
Who is the sister of for king and country?
=================================================== Answers ==================================================== 
['Rebecca St. James']
================================================= Generation 1 ================================================= 
Rebecca St. James

### Explanation:
Rebecca St. James is the sister of for king and country. She is a Christian recording artist and speaker. The Smallbones are the younger brothers of Rebecca St. James, and the brothers-in-law of Jacob Fink, former bassist for the band Foster the People. They were raised with another sister, Libby, and three other brothers, Ben, Dan, and Josh. On
=================================================== Output 1 ===================================================
RAG Label: 1
================================================= Generation 2 ================================================= 
Courtney Helm

### Explanation:
Courtney Helm is the sister of for king and country.
=================================================== Output 2 ===================================================
RAG Label: 0
\end{lstlisting}
\end{tcolorbox}
\caption{Examples of RAG labels for free-form generation.}
\label{fig:raglabel1}
\end{figure*}

\begin{figure*}[ht]
\begin{tcolorbox}[left=1mm,right=1mm,top=0.mm, bottom=0mm,colback=white]
\begin{lstlisting}[style=demo, breaklines=true, breakindent=0pt]
================================================== Question ==================================================== 
Study: Vaccine for Breast, Ovarian Cancer Has Potential.
=================================================== Answers ==================================================== 
True
================================================= Generation 1 ================================================= 
True

### Explanation:
The study is about a vaccine for breast and ovarian cancer. The vaccine is
=================================================== Output 1 ===================================================
RAG Label: 1
================================================= Generation 2 ================================================= 
False

### Explanation:
The statement is false because the vaccine is not for breast and ovarian cancer.
=================================================== Output 2 ===================================================
RAG Label: 0
\end{lstlisting}
\end{tcolorbox}
\caption{Examples of RAG labels for close-set generation.}
\label{fig:raglabel2}
\end{figure*}

\section{Prompt Formats}
\label{appendix:prompt}

We demonstrate our prompts for different tasks such as OpenQA, fact-checking, and multi-choice reasoning in Figures~\ref{fig:prompt-qa}, \ref{fig:prompt-factcheck}, and \ref{fig:prompt-choice}, respectively.

\begin{figure*}[ht]
\begin{tcolorbox}[left=1mm,right=1mm,top=0.mm, bottom=0mm,colback=white]
\begin{lstlisting}[style=demo, breaklines=true, breakindent=0pt]
=============================================== Prompt (OpenQA) ================================================ 
### Paragraph:
[1] For King & Country (band)
Their father was a music promoter, and Joel recalls, "going to these rock concerts, sitting on my father\'s shoulders, plugging my ears...Honestly, in a lot of ways I feel like music chose me and as I grew older, I made a clear decision to fully lean into it." The Smallbones are the younger brothers of Christian recording artist and speaker Rebecca St. James, and the brothers-in-law of Jacob Fink, former bassist for the band Foster the People. They were raised with another sister, Libby, and three other brothers, Ben, Dan, and Josh. On ...

### Instruction: 
Who is the sister of for king and country?

### Response: 
================================================= Continuation ================================================= 
Rebecca St. James
================================================== Generation ================================================== 
Rebecca St. James

### Explanation:
Rebecca St. James is the sister of for king and country. She is a Christian recording artist and speaker. The Smallbones are the younger brothers of Rebecca St. James, and the brothers-in-law of Jacob Fink, former bassist for the band Foster the People. They were raised with another sister, Libby, and three other brothers, Ben, Dan, and Josh. On
==================================================== Output ====================================================
RAG Score: P(`Rebecca St. James'|prompt) = 0.595
RAG Label: 1
\end{lstlisting}
\end{tcolorbox}
\caption{Example prompt and outcomes of each step for NQ and TQA dataset.}
\label{fig:prompt-qa}
\end{figure*}

\begin{figure*}[ht]
\begin{tcolorbox}[left=1mm,right=1mm,top=0.mm, bottom=0mm,colback=white]
\begin{lstlisting}[style=demo, breaklines=true, breakindent=0pt]
============================================ Prompt (Fact Checking) ============================================ 
Below is an instruction that describes a task. Write a response that appropriately completes the request.

### Paragraph:
[1] Gustav Gaudernack
potential of dendritic cells (DCs) and in 2005, Gaudernack's group published results from a phase I/II clinical trial in prostate cancer patients using autologous DCs loaded with tumor mRNA as a vaccine. This study demonstrated that vaccination with autologous DCs transfected with mRNA derived from three prostate cancer cell lines was safe and an improved clinical outcome was significantly related to immune responses against the vaccine. Furthermore, Gaudernack and colleagues initiated a phase I/II clinical trial for treatment of malignant melanoma with autologous tumor-mRNA transfected DC vaccines. These data clearly demonstrated vaccine-specific immune responses with a broad specter of ...

### Instruction:
Is the following statement correct or not? Say true if it's correct; otherwise say false.

### Input:
Study: Vaccine for Breast, Ovarian Cancer Has Potential

### Response:
================================================= Continuation ================================================= 
True
================================================== Generation ================================================== 
true

### Explanation:
The study is about a vaccine for breast and ovarian cancer. The study has ...
==================================================== Output ====================================================
P(`true' |prompt) = 0.116
P(`false'|prompt) = 0.109
RAG Label: 1
\end{lstlisting}
\end{tcolorbox}
\caption{Example prompt and outcomes of each step for the Pubhealth dataset.}
\label{fig:prompt-factcheck}
\end{figure*}
% \FloatBarrier

\begin{figure*}[th]
\begin{tcolorbox}[left=1mm,right=1mm,top=0.mm,bottom=0mm,colback=white]
\begin{lstlisting}[style=demo, breaklines=true, breakindent=0pt]
======================================= Prompt (Multi-choice Reasoning) ======================================== 
Below is an instruction that describes a task. Write a response that appropriately completes the request.

### Paragraph:
[1] Rheumatic fever
Rheumatic fever may occur following an infection of the throat by the bacterium "Streptococcus pyogenes". If the infection is untreated rheumatic fever can occur in up to three percent of people. The underlying mechanism is believed to involve the production of antibodies against a person\'s own tissues. Due to their genetics, some people are more likely to get the disease when exposed to the bacteria than others. Other risk factors include malnutrition and poverty. Diagnosis of RF is often based on the presence of signs and symptoms in combination with evidence of a recent streptococcal infection. Treating people who have strep ...

### Instruction:
Given four answer candidates, A, B, C and D, choose the best answer choice.

### Input:
Which factor will most likely cause a person to develop a fever?
A: a leg muscle relaxing after exercise
B: a bacterial population in the bloodstream
C: several viral particles on the skin
D: carbohydrates being digested in the stomach

### Response: 
================================================= Continuation ================================================= 
B
================================================== Generation ================================================== 
B

### Explanation:
The bacteria Streptococcus pyogenes is a common cause of throat
==================================================== Output ====================================================
P(`A'|prompt) = 0.121
P(`B'|prompt) = 0.309
P(`C'|prompt) = 0.061
P(`D'|prompt) = 0.100
RAG Label: 1

\end{lstlisting}
\end{tcolorbox}
\caption{Example prompt and outcomes of each step for the ARC-Challenge dataset.}
\label{fig:prompt-choice}
\end{figure*}
\FloatBarrier

%%%%%%%%%%%%%%%%%%%%%%%%%%%%%%%%%%%%%%%%%%%%%%%%%%%%%%%%%%%%%%%%%%%%%%%%%%%%%%%
%%%%%%%%%%%%%%%%%%%%%%%%%%%%%%%%%%%%%%%%%%%%%%%%%%%%%%%%%%%%%%%%%%%%%%%%%%%%%%%

\end{document}